\documentclass{article}

\usepackage[preprint]{neurips_2026}
\usepackage{microtype}
\usepackage{graphicx}
\usepackage{subcaption}
\usepackage{booktabs} % for professional tables

\usepackage{algorithm2e}
\usepackage[noend]{algpseudocode}
\usepackage{listings}

\lstdefinestyle{mystyle}{
    backgroundcolor=\color{backcolour},   
    commentstyle=\color{codegreen},
    keywordstyle=\color{magenta},
    numberstyle=\tiny\color{codegray},
    stringstyle=\color{codepurple},
    basicstyle=\ttfamily\footnotesize,
    breaklines=true,                 
    captionpos=b,                    
    keepspaces=true,                 
    showspaces=false,                
    showstringspaces=false,
    showtabs=false,                  
    tabsize=2
}
\usepackage{hyperref}

\usepackage[utf8]{inputenc} % allow utf-8 input
\usepackage[T1]{fontenc}    % use 8-bit T1 fontsd
\usepackage{hyperref}       % hyperlinks
\usepackage{url}            % simple URL typesetting
\usepackage{booktabs}       % professional-quality tables
\usepackage{amsfonts}       % blackboard math symbols
\usepackage{nicefrac}       % compact symbols for 1/2, etc.
\usepackage{microtype}      % microtypography
\usepackage{xcolor}         % colors
\usepackage{amsthm, amssymb, amsfonts, latexsym, mathtools}
\usepackage{comment}
\usepackage{here}
\usepackage{caption}
\usepackage{subcaption}
\usepackage{wrapfig}
\usepackage{multirow}
\usepackage{color}
\usepackage{wrapfig}
\usepackage{threeparttable}
\usepackage{listings}
\usepackage{xcolor} 
\usepackage{enumitem}
\usepackage{array}

\usepackage{pifont}

\definecolor{codegreen}{rgb}{0,0.6,0}
\definecolor{codegray}{rgb}{0.5,0.5,0.5}
\definecolor{codepurple}{rgb}{0.58,0,0.82}
\definecolor{backcolour}{rgb}{0.95,0.95,0.95}

\usepackage{booktabs}

\usepackage[utf8]{inputenc} % allow utf-8 input
\usepackage[T1]{fontenc}    % use 8-bit T1 fonts
\usepackage{hyperref}       % hyperlinks
\usepackage{url}            % simple URL typesetting
\usepackage{booktabs}       % professional-quality tables
\usepackage{amsfonts}       % blackboard math symbols
\usepackage{nicefrac}       % compact symbols for 1/2, etc.
\usepackage{microtype}      % microtypography
\usepackage{xcolor}         % colors

\usepackage{amsmath,amsfonts,bm}

\def\eqref#1{equation~\ref{#1}}
\def\1{\bm{1}}

\def\mH{{\bm{H}}}

\def\mW{{\bm{W}}}
\def\mX{{\bm{X}}}
\def\mY{{\bm{Y}}}
\def\mZ{{\bm{Z}}}

\DeclareMathAlphabet{\mathsfit}{\encodingdefault}{\sfdefault}{m}{sl}
\SetMathAlphabet{\mathsfit}{bold}{\encodingdefault}{\sfdefault}{bx}{n}

\title{SiamJEPA: On the Role of Siamese Student Encoders in JEPA}

\author{%
  Makoto Yamada\\
  Okinawa Institute of Science and Technology\\
  \texttt{makoto.yamada@oist.jp} \\
}

\begin{document}

\maketitle

\begin{abstract}
Joint Embedding Predictive Architectures (JEPAs) have emerged as a promising framework for self-supervised representation learning by predicting latent embeddings of masked regions rather than reconstructing pixels. Existing JEPA methods typically employ a single student encoder, leaving the role of Siamese student encoders largely unexplored. In this paper, we propose Siamese JEPA (SiamJEPA), a JEPA framework with masked Siamese student encoders and an exponential moving average (EMA) teacher, which can also be viewed as a JEPA formulation of the brain-inspired representation learning model PhiNet. We further introduce Random Shuffle Teacher (RST), which removes spatial correspondence in teacher targets to encourage semantic patch representations, and develop an RST-based semantic-to-spatial curriculum. Experiments on ImageNet show that Siamese student encoders effectively regularize the JEPA objective, improving representation separability and accelerating early-stage learning. Moreover, under RST, stronger Siamese regularization substantially increases class-discriminative information in individual patch tokens, suggesting that the semantic bias arises from the interaction between RST and the Siamese objective rather than from shuffling alone. SiamJEPA consistently outperforms comparable single-encoder JEPA variants under limited training budgets. With RST-based curriculum learning and a ViT-Base backbone, SiamJEPA achieves 73.3\% linear-probing accuracy after 450 epochs using a substantially simpler masking strategy, compared with 72.9\% for I-JEPA after 600 epochs. Under I-JEPA's linear-probing protocol, the same checkpoint achieves 74.2\%, outperforming I-JEPA (72.9\%) and DSeq-JEPA (73.5\%). These results demonstrate that Siamese student encoders provide an effective inductive bias for predictive representation learning and can be further enhanced through semantic-to-spatial curriculum learning. The source code is publicly
available at \url{https://github.com/oist/SiamJEPA}.
\end{abstract}

\section{Introduction}
Recently, representation learning has become a fundamental technique in computer vision, natural language processing, and robotics. Among various approaches, self-supervised learning (SSL) has emerged as one of the most successful paradigms for learning transferable representations from large-scale unlabeled data. Conceptually, several early studies introduced the principles of modern SSL, including information maximization (IMAX) and predictability maximization (PMAX) \citep{becker1992self,zemel1990discovering,schmidhuber1993discovering}. More recently, advances in deep neural networks, particularly Transformer architectures such as the Transformer \citep{vaswani2017attention} and the Vision Transformer (ViT) \citep{dosovitskiy2021an}, have enabled highly effective self-supervised representation learning methods.

A major family of SSL methods is based on joint embedding architectures, which learn representations by mapping different views of the same input into a shared latent space using deep neural networks. Representative examples include SimCLR \citep{chen2020simple}, BYOL \citep{grill2020bootstrap}, SimSiam \citep{chen2021exploring}, and DINO \citep{caron2021emerging,oquab2024dinov,simeoni2025dinov3}. A central challenge of joint embedding methods is to prevent representation collapse. To address this issue, various techniques have been proposed, including contrastive learning with data augmentation \citep{chen2020simple}, stop-gradient operations, and exponential moving average (EMA) teacher networks \citep{grill2020bootstrap,chen2021exploring}. Together with large-scale engineering efforts \citep{oquab2024dinov,simeoni2025dinov3}, these advances have established SSL as a cornerstone for training modern vision foundation models.

More recently, Joint Embedding Predictive Architectures (JEPAs) have emerged as a promising extension of joint embedding methods. Instead of directly aligning latent representations, JEPAs learn to predict masked latent representations from visible context. Representative examples include I-JEPA \citep{assran2023self} and V-JEPA \citep{bardes2024revisiting,assran2025v}. One of the key challenges in JEPA is preventing representation collapse while preserving informative latent representations. To this end, LeJEPA introduces Sketched Isotropic Gaussian Regularization (SIGReg) \citep{balestriero2025lejepa}, which regularizes the embedding space toward an isotropic Gaussian distribution and improves training stability. In contrast, most existing JEPA models, including I-JEPA and V-JEPA, follow the student--teacher paradigm of BYOL and DINO, relying on stop-gradient operations and exponential moving average (EMA) updates to avoid representation collapse.

Independently, brain-inspired representation learning methods such as PhiNets \citep{ishikawa2025phinets} have been proposed. PhiNets is motivated by the biological circuitry of the hippocampus and neocortex, drawing inspiration from the temporal predictive hypothesis \citep{chen2022predictive} and the Complementary Learning Systems (CLS) theory \citep{mcclelland1995there}. Specifically, PhiNets employ Siamese \emph{student} encoders together with an EMA-based target network. Within this framework, the Siamese student encoders are intended to model the temporal predictive hypothesis, while the student--teacher learning mechanism based on exponential moving average (EMA) can be interpreted as a computational analogue of fast and slow learning in the CLS theory. More recently, Transformer-based extensions of PhiNets have also been proposed \citep{yamada2025phinet}, which learn to predict the next frame in latent space and can be regarded as a JEPA method.

These approaches share architectural similarities with recent masked prediction methods, including SiamMAE \citep{gupta2023siamese}, CropMAE \citep{cropmae}, and RSP \citep{jang2024visual}, all of which employ Siamese encoders to learn from video data. However, unlike JEPA-based methods, these approaches are primarily designed for pixel- or feature-level reconstruction objectives rather than latent-space prediction. Although they have demonstrated strong performance on dense prediction tasks such as segmentation and pose estimation, their effectiveness for learning general-purpose visual representations remains less understood. More importantly, despite the increasing adoption of Siamese encoders in recent self-supervised learning methods, their role within latent predictive architectures such as JEPA has not been systematically investigated. 

\begin{center}
\textit{What role do Siamese student encoders play in Joint Embedding Predictive Architectures (JEPAs)?}
\end{center}

In this paper, we investigate the role of Siamese student encoders in a JEPA framework. To this end, we propose SiamJEPA, a masked latent prediction architecture based on Siamese student encoders. Unlike existing JEPA models, SiamJEPA employs two independently masked views and learns to predict the latent representations of masked tokens from their unmasked counterparts without relying on pixel reconstruction. Furthermore, the proposed framework introduces a regularization parameter that controls the influence of the Siamese student encoders, with the conventional single-encoder JEPA recovered as a special case. Moreover, to efficiently train the SiamJEPA model, we propose the random shuffle teacher (RST) and the semantic-to-spatial curriculum learning. We evaluate SiamJEPA on the ImageNet linear-probing benchmark with a ViT-Base backbone \citep{dosovitskiy2021an}, comparing against JEPA-like methods and Masked Autoencoders (MAE) \citep{he2022masked}.

Our contributions are summarized as follows:
\begin{itemize}
\item We propose SiamJEPA, a JEPA framework with Siamese student encoders.
\item We introduce the Random Shuffle Teacher (RST) and a semantic-to-spatial curriculum that first learns spatially invariant semantic representations and restores spatial correspondence.
\item We provide the first systematic study of Siamese student encoders in JEPA, showing that they effectively regularize the representation space and accelerate convergence over single-encoder JEPA.
\item SiamJEPA is highly training-efficient, matching MAE with fewer epochs and achieving 74.2\% ImageNet linear-probing accuracy under I-JEPA's protocol, outperforming I-JEPA (72.9\%) and DSeq-JEPA (73.5\%) with fewer training epochs (450 vs.\ 600).
\end{itemize}

\begin{figure*}[t!]
    \centering
      \begin{subfigure}[t]{0.45\textwidth}
        \centering
        \includegraphics[width=.99\textwidth]{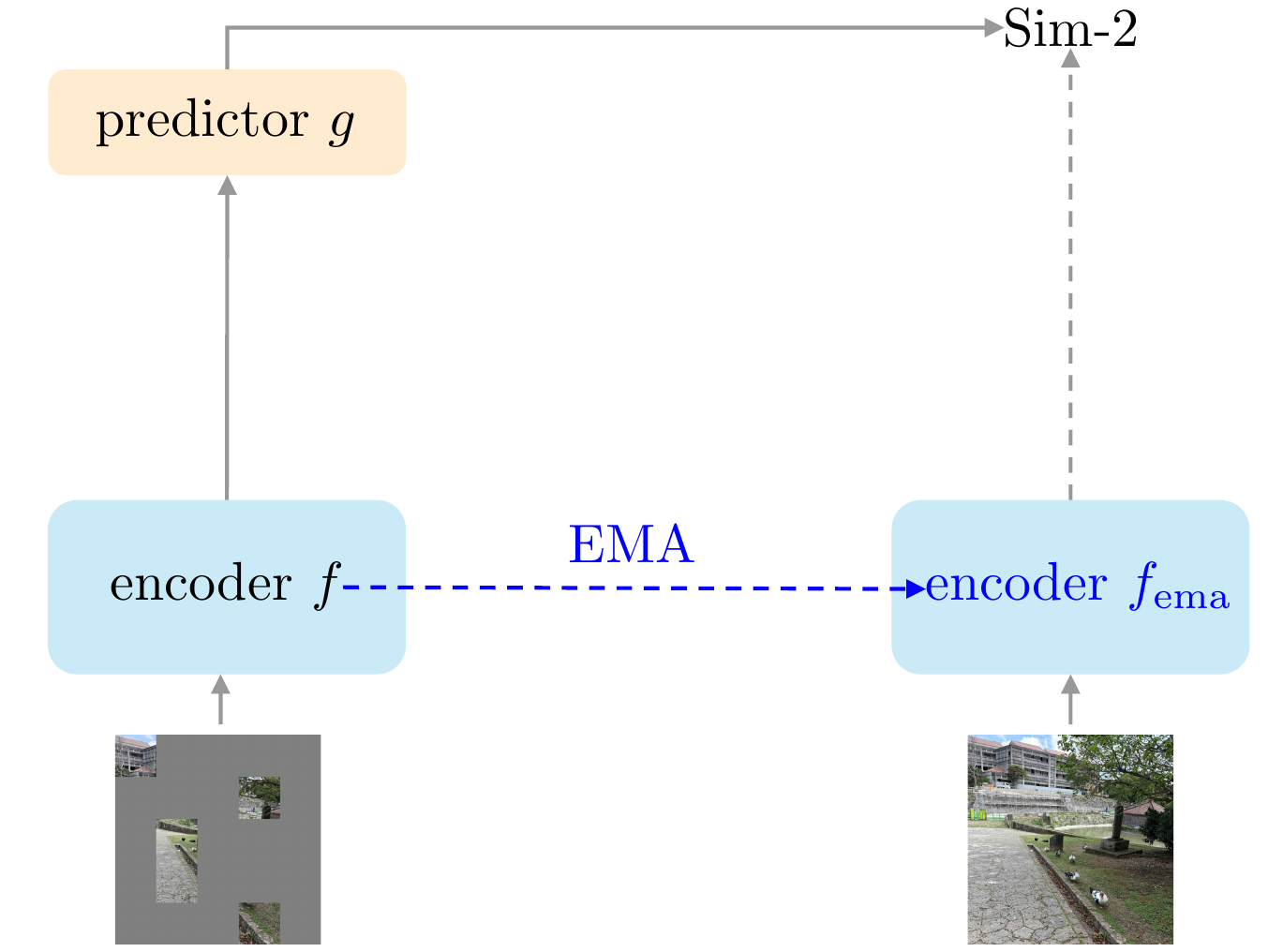}%
        \caption{A JEPA architecture. \label{subfig:arch-jepa}}
    \end{subfigure}
        \begin{subfigure}[t]{0.45\textwidth}
        \centering
        \includegraphics[width=.99\textwidth]{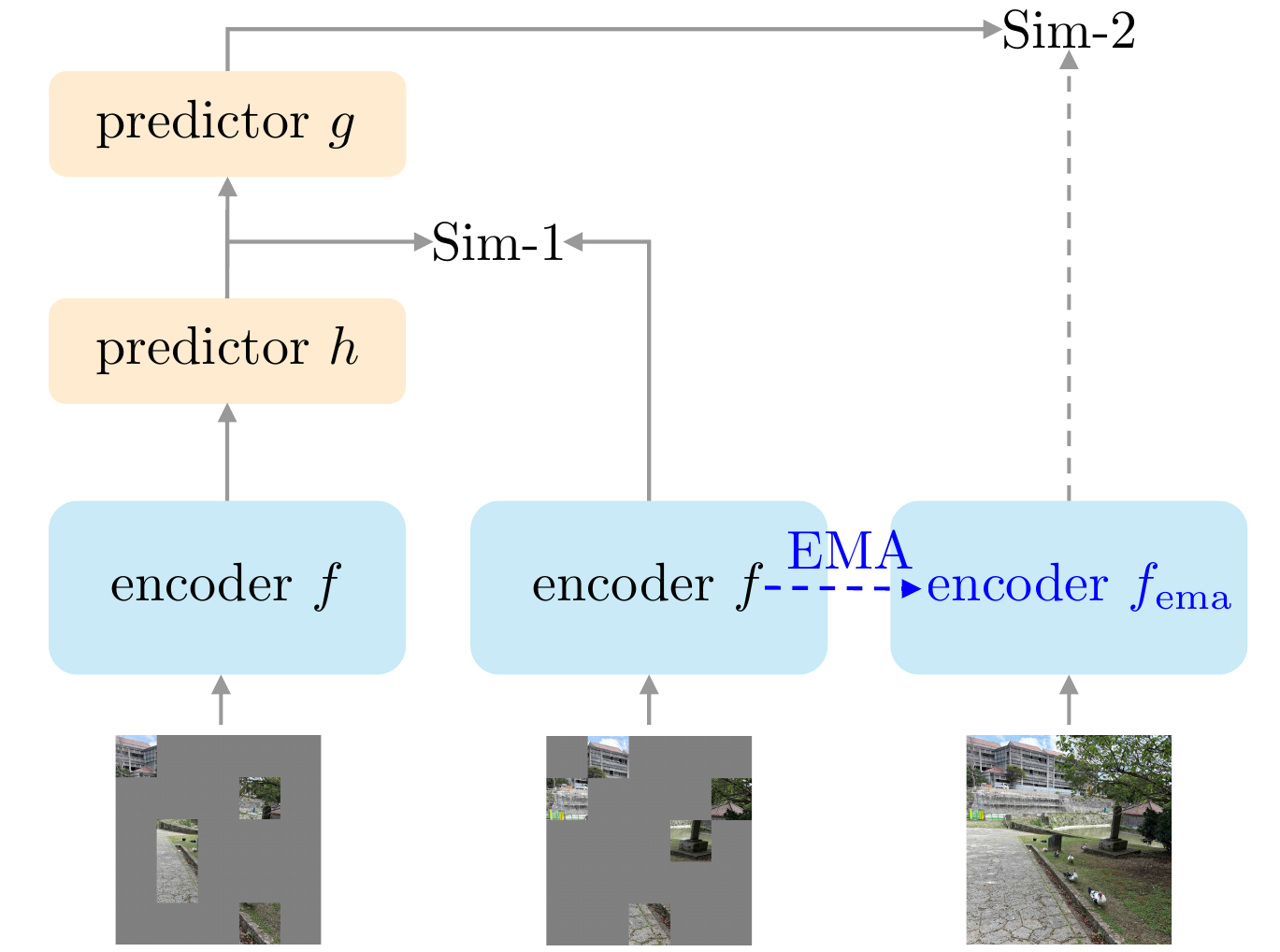}%
        \caption{SiamJEPA architecture (a.k.a., Phinet architecture \citep{yamada2025phinet}). \label{subfig:arch-rsp}}
    \end{subfigure}
    \caption{JEPA and SiamJEPA architectures. Sim-1 is a loss function to align the Siamese encoders. The dashed line represents the StopGradient operator. In our implementation, the two masking sets are disjoint.  Note that the SiamJEPA architecture is inspired by the brain-inspired representation learning called PhiNet. }
    \label{fig:arch}
    \vspace{-.1in}
\end{figure*}

\section{Related Work}
This section reviews Joint Embedding Predictive Architectures (JEPAs). Due to space limitations, we provide an extended discussion of related work on joint-embedding architectures and masked modeling in the Appendix. 

\noindent {\bf Masked modeling without pixel reconstruction:} Building on this idea, JEPAs were proposed to predict masked representations directly in latent space rather than reconstructing pixels \citep{lecun2022path}, and have since been extended from images (I-JEPA) \citep{assran2023self} to videos (V-JEPA) \citep{assran2025v} and, more recently, world-model learning via V-JEPA 2. Both I-JEPA and V-JEPA use an exponential moving average (EMA) teacher network to prevent collapse. LeJEPA \citep{balestriero2025lejepa} instead introduces Sketched Isotropic Gaussian Regularization (SIGReg), encouraging encoder outputs toward an isotropic Gaussian by minimizing a sliced Wasserstein distance to a Gaussian reference. Recently, \citet{wu2026visreg} proposed VISReg, an alternative to SIGReg that separately regularizes the center, scale, and shape of the representation distribution. Latent prediction's key advantage is that it focuses on semantic information, often learning useful representations more efficiently than pixel reconstruction. 

PhiNetv2 \citep{yamada2025phinet} was proposed as a latent-prediction counterpart of Siamese masked architectures, predicting future latent representations rather than pixels and showing strong video representation-learning performance. Several questions remain open, however: its effectiveness for image representation learning is largely unexplored, Siamese encoders have received relatively little attention in JEPA-style architectures, and existing masked representation learning methods generally assume carefully designed context masking is essential. SiamJEPA is intrinsically a masked PhiNetv2 model, and this paper investigates the role of Siamese student encoders in a JEPA. 

The idea behind JEPA is rooted in both classical and modern self-supervised learning. Earlier frameworks such as Information Maximization (IMAX) \citep{zemel1990discovering} and its predictive counterpart, Predictability Maximization (PMAX) \citep{schmidhuber1993discovering}, share JEPA's core principle that useful representations should capture information predictable from related observations. JEPA can thus be viewed as a modern realization of these ideas, enabled by transformer architectures and large-scale optimization.

\noindent {\bf Curriculum learning for JEPA:} DSeq-JEPA \citep{he2025dseq} introduces a curriculum-like progression by sequentially predicting regions in order of visual importance; our curriculum instead operates across training stages, first encouraging spatially invariant semantic representations via RST, then restoring spatial correspondence to refine them. DSeq-JEPA builds on single-encoder I-JEPA with a curriculum over region order, whereas ours introduces a semantic-to-spatial curriculum within a Siamese JEPA -- the two approaches are complementary rather than mutually exclusive. In principle, the region-level curriculum of DSeq-JEPA could be combined with our semantic-to-spatial curriculum, potentially leading to further improvements in representation quality.

\section{JEPA with Siamese encoders (SiamJEPA)}

In this section, we propose the SiamJEPA model, which is inspired by the brain-inspired representation learning method called PhiNet architecture with Transformer encoder \citep{yamada2025phinet}. Figure \ref{fig:arch} shows the architecture of both JEPA and SiamJEPA architectures. The key difference from the I-JEPA model is the existence of an additional encoder in the student network.

\subsection{Siamese student encoders}
We employ Siamese student encoders with masking augmentation applied independently to each encoder. Let $\mX \in \mathbb{R}^{m \times d_{\textnormal{in}}}$ denote the patchified input image and  $m$ is the number of image patches. The outputs of the student and teacher vision transformer encoders are given by
\begin{align*}
\mH^{(1)} &= f(\textnormal{Mask}(\mX, M_1)),~\mH^{(2)} = f(\textnormal{Mask}(\mX, M_2)),~\mY = f_{\textnormal{ema}}(\mX),
\end{align*}
where $\mH^{(1)}\in \mathbb{R}^{N \times (m_1+1) \times d}, \mH^{(2)}\in \mathbb{R}^{N \times (m_2+1) \times d}, \mY \in \mathbb{R}^{N \times (m+1) \times d}$ are the output representations, $N$ is the batch size, $m_1=|M_1|$ and $m_2=|M_2|$ are the number of unmasked tokens, $M_1$ and $M_2$ are the masking indices, and $d$ is the embedding dimension. Here, $f(\cdot)$ denotes the Siamese student encoder with shared weights, while $f_{\textnormal{ema}}(\cdot)$ denotes the teacher encoder whose parameters are updated using an exponential moving average (EMA). The function $\textnormal{Mask}(\mX,M)$ denotes the masking operator. In our implementation, the two masking sets are disjoint, i.e., $M_1 \cap M_2 = \emptyset.$ This non-overlapping masking strategy is crucial for preventing shortcut learning in SiamJEPA. Positional embeddings are added to the input patches before they are fed into the encoders.

We further decompose the encoder outputs as $\mH^{(1)}
=
[\mH_{\textnormal{cls}}^{(1)};\mH_{\textnormal{patch}}^{(1)}],
\mH^{(2)}
=
[\mH_{\textnormal{cls}}^{(2)};\mH_{\textnormal{patch}}^{(2)}],
\mY =
[\mY_{\textnormal{cls}};\mY_{\textnormal{patch}}]$, where $[\cdot,;\cdot]$ denotes token-wise concatenation, $\mH_{\textnormal{cls}} \in \mathbb{R}^{N \times 1 \times d}$ denotes the CLS token representation, and $\mH_{\textnormal{patch}} \in \mathbb{R}^{N \times m \times d}$ denotes the patch token representations.

\subsection{Predictor networks}
We employ two predictor networks, denoted by $h(\cdot)$ and $g(\cdot)$. The predictor $h(\cdot)$ is responsible for aligning the global representations produced by the Siamese student encoders, whereas the predictor $g$ predicts the latent representations of the masked tokens. This design follows the key idea of the PhiNet architecture \citep{ishikawa2025phinets,yamada2025phinet}.

For the predictor $h(\cdot)$, we employ the linear model as $h(\mH) = \mH \mW$, where $\mW \in \mathbb{R}^{d\times d}$ is a linear transform to help aligning the two Siamese student encoders.

For the predictor $g(\cdot)$, we adopt the probabilistic predictor architecture of \citet{jang2024visual}, which is inspired by  \citet{pmlr-v80-denton18a} and \citet{hafner2020dreamerv2}. Specifically, we define the posterior and prior distributions as
\begin{align*}
\textnormal{Posterior:}\quad
&\mZ^{(1)} \sim q(\mZ^{(1)} \mid h(\mH^{(1)}_{\textnormal{cls}}), \mH^{(2)}_{\textnormal{cls}}),~~\textnormal{Prior:}\quad
\widehat{\mZ}^{(1)} \sim p(\widehat{\mZ}^{(1)} \mid h(\mH^{(1)}_{\textnormal{cls}})).
\end{align*}
The posterior incorporates information from both Siamese branches, whereas the prior is conditioned only on the representation of the first branch. We also denote the flipped version of $h(\mH^{(1)}_{\textnormal{cls}})$ and $\mH^{(2)}_{\textnormal{cls}}$ as $\mZ^{(2)}$ and $\widehat{\mZ}^{(2)}$, respectively. Consequently, the latent variable $\mZ$ models the uncertainty of the latent representation given the available context. $q(\cdot)$ and $p(\cdot)$ denote the posterior and prior distributions, respectively, both parameterized by two-layer neural networks. Note that before constructing these distributions, we first apply a projector head with batch normalization. 

For the predictor $g(\cdot)$, we employ a transformer model as
\begin{align*}
\widehat{\mY}^{(1)} = g(h(\mH^{(1)}),\mZ^{(1)}) \in \mathbb{R}^{N\times (m+1) \times d}~~\textnormal{and}~~\widehat{\mY}^{(2)} = g(h(\mH^{(2)}),\mZ^{(2)}) \in \mathbb{R}^{N\times (m+1) \times d}, 
\end{align*}
where $\widehat{\mY}^{(1)} = [\widehat{\mY}^{(1)}_{\textnormal{cls}};\widehat{\mY}^{(1)}_{\textnormal{patch}}]\in \mathbb{R}^{N \times (m+1) \times d}$ and $\widehat{\mY}^{(2)} = [\widehat{\mY}^{(2)}_{\textnormal{cls}};\widehat{\mY}^{(2)}_{\textnormal{patch}}]\in \mathbb{R}^{N \times (m+1) \times d}$. 

\subsection{Loss functions}
We optimize two complementary objectives: (Sim-1) aligning the global representations of the Siamese student encoders, and (Sim-2) predicting the latent representations of the masked tokens. Specifically, Sim-1 is optimized using the Kullback--Leibler (KL) divergence, whereas Sim-2 is optimized using the normalized mean squared error (NMSE).

\noindent {\bf Sim-1:}
\begin{align*}
\textnormal{KL}^{(1)} &= \textnormal{KL}(q(\mZ | [h(\mH^{(1)}_{\textnormal{cls}}),\mH^{(2)}_{\textnormal{cls}}])\|p(\mZ|h(\mH^{(1)}_{\textnormal{cls}}))),\\
\textnormal{KL}^{(2)} &= \textnormal{KL}(q(\mZ | [h(\mH^{(2)}_{\textnormal{cls}}),\mH^{(1)}_{\textnormal{cls}}])\|p(\mZ | h(\mH^{(2)}_{\textnormal{cls}}))).
\end{align*}

The posterior distribution is conditioned on both views, whereas the prior distribution is conditioned on only one view. Therefore, minimizing the KL divergence encourages the prior distribution inferred from a single view to approximate the posterior distribution inferred from both views. Consequently, the latent variable obtained from one view becomes predictive of the latent representation that would otherwise require information from both views, thereby encouraging the encoder to learn representations that are robust to view variations.

We also found that stopping the gradient through the prior branch consistently improves optimization stability and downstream performance. Therefore, we employ the following objective:
\begin{align*}
\textnormal{KL}^{(1)}_{\textnormal{sg}} &= \textnormal{KL}(q(\mZ | [h(\mH^{(1)}_{\textnormal{cls}}),\mH^{(2)}_{\textnormal{cls}}])\|\textnormal{sg}(p(\mZ|h(\mH^{(1)}_{\textnormal{cls}})))),\\
\textnormal{KL}^{(2)}_{\textnormal{sg}} &= \textnormal{KL}(q(\mZ | [h(\mH^{(2)}_{\textnormal{cls}}),\mH^{(1)}_{\textnormal{cls}}])\|\textnormal{sg}(p(\mZ | h(\mH^{(2)}_{\textnormal{cls}})))),
\end{align*}
where $\textnormal{sg}(\cdot)$ is the stopgradient operator. This heuristic was originally introduced empirically by \citet{yamada2025phinet}.

This probabilistic formulation was originally introduced for future-frame prediction \citep{pmlr-v80-denton18a}. In this work, we adopt the same principle to encourage consistency between two augmented views. By requiring the prior inferred from a single view to match the posterior inferred from both views, the encoder is encouraged to capture information that is shared across views while discarding view-specific variations. This complements the representation learning objective of SiamJEPA and promotes more invariant latent representations.

Following PhiNet v2 \citep{yamada2025phinet}, we adopt the KL divergence for the Siamese student encoder. Although other objectives, such as cosine similarity or mean squared error in latent space, could also be considered, the probabilistic formulation provides an intuitive interpretation of latent consistency and demonstrated strong empirical performance in our preliminary experiments.

\noindent {\bf Sim-2:}
\begin{align}
\textnormal{MSE}^{(\ell)} &= \frac{1}{|\bar{M}|}\|\textnormal{Mask}(\mY^{(\ell)}_{\textnormal{patch}} - \widehat{\mY}^{(\ell)}_{\textnormal{patch}},\bar{M})\|_{\textnormal{Frob}}^2,~\ell \in \{1,2\},
\end{align}
where $\|\cdot\|_{\textnormal{Frob}}$ is the Frobenius norm and $\bar{M} = M \setminus (M_1 \cup M_2)$ and $M = \{1,2,\ldots, m\}$ as the set of index of patches. 

Final loss function is given as
\begin{align*}
L = \frac{1}{2}(\textnormal{MSE}^{(1)} + \textnormal{MSE}^{(2)}) + \frac{\lambda_{\textnormal{KL}}}{2}(\textnormal{KL}^{(1)}_{\textnormal{sg}} + \textnormal{KL}^{(2)}_{\textnormal{sg}}).
\end{align*}
Note that if we set $\lambda_{\textnormal{KL}} = 0$, the SiamJEPA model is similar to that of single encoder JEPA model. Since our model uses the special decoder depends on the posterior distribution, we name the SiamJEPA model with $\lambda_{\textnormal{KL}} = 10^{-5}$ as a JEPA-like method. 

\subsection{Random Shuffle Teacher (RST)}
\label{sec:rst-method}

A key objective of JEPA is to align student and teacher representations at the patch level. While this patch-wise prediction encourages the model to capture spatially structured information, it may also allow the prediction task to rely heavily on spatial correspondence between student and teacher tokens. To encourage each patch representation to capture more global semantic information, we propose the \emph{Random Shuffle Teacher} (RST), which randomly permutes the teacher patch tokens before computing the prediction objective.

The key intuition behind RST is that random permutation breaks the direct spatial correspondence between student and teacher tokens. Consequently, the student cannot reliably predict the target representation based solely on positional or local spatial information. To minimize the prediction objective under this perturbation, the model is encouraged to encode semantic information that is shared across patches, rather than relying primarily on their spatial locations. In this way, RST acts as a simple regularization mechanism that promotes the distribution of global semantic information across individual patch representations.

Building on this observation, we further propose an RST-based curriculum learning strategy. Specifically, we first train the model with RST to encourage the emergence of globally semantic patch representations. In the later stages of training, we remove the random shuffling and restore the original spatial correspondence between student and teacher tokens. This allows the model to subsequently refine spatially structured representations while retaining the semantic representations acquired during the initial RST stage.

\begin{table}[t]
\centering
\caption{ImageNet linear probing performance comparison. The results demonstrate that enforcing consistency between the Siamese student encoders improves the quality of the learned representations, resulting in higher ImageNet linear probing performance.  Note that MAE is evaluated using the CLS token, whereas SiamJEPA is evaluated using mean pooling. $^\dagger$ We conduct the experiments using the official implementation available at \url{https://github.com/facebookresearch/mae}. Results for the original I-JEPA and Dseq-JEPA are taken from \citep{assran2023self} and \citep{he2025dseq} (600 training epochs), respectively. $^\ddagger$ Same checkpoint as the row above (total epoch 450), evaluated with the I-JEPA linear-probing protocol (avg-pooled/4-layer-concat features, best of a 12-head grid) instead of 12th-layer mean pooling.}
\vspace{.05in}
\label{tab:performance_all}
\begin{tabular}{lcc}
\toprule
\textbf{Method}  &Epochs&\textbf{Top-1 Acc. (\%)}  \\
\midrule
MAE$^\dagger$  & 400 & 61.9  \\
MAE  \citep{he2022masked}&  1600 &68.0  \\
CAE \citep{chen2024context} & 1600 & 70.4 \\
I-JEPA \citep{assran2023self} & 600 & 72.9 \\
Dseq-JEPA \citep{he2025dseq} & 600 & 73.5 \\
\midrule
\multirow{1}{*}{SiamJEPA with RST (12th layer)}  &  450 & 73.3  \\
\multirow{1}{*}{{SiamJEPA with RST (I-JEPA protocol)}}  &  450$^\ddagger$ & {\bf 74.2}  \\
\bottomrule
\end{tabular}
\vspace{-.2in}
\end{table}

\section{Experiments}
In this section, we conduct the ablation study of SiamJEPA and compared its performance with the masked autoencoder (MAE) \citep{he2022masked}, context autoencoder \citep{chen2024context},  I-JEPA \citep{assran2023self}, and Dseq-JEPA \citep{he2025dseq}. The main purpose of this experiment is not to achieve the state of the art performance in JEPA.  

\subsection{Setup and implementation details}
All ablation studies are conducted under the same experimental setting (Table~\ref{tab:config}). We employ an EMA momentum schedule that increases from 0.99 during epochs 1--200, to 0.999 during epochs 201--300, and to 0.9999 during epochs 301--400.

All experiments are performed using NVIDIA V100, A100, or H100 GPUs. Based on the findings from these studies, we select the hyperparameters and train the final models for 400 pre-training epochs for the main comparisons. 

Note that our implementation is built upon the official MAE codebase \footnote{\url{https://github.com/facebookresearch/mae}} rather than the official I-JEPA implementation. Consequently, although the training dynamics differ from those reported for I-JEPA, our implementation provides a common experimental framework for comparing reconstruction-based and JEPA-based methods.

\subsection{Comparison to other SSL methods}
 As shown in Table~\ref{tab:performance_all}, SiamJEPA achieves performance comparable to that of MAE and CAE while requiring substantially fewer training epochs. In particular, SiamJEPA outperforms MAE using less than one-quarter of the training epochs and achieves performance comparable to CAE, demonstrating significantly improved training efficiency.

Compared with I-JEPA \citep{assran2023self} and DSeq-JEPA \citep{he2025dseq}, SiamJEPA without curriculum learning achieves lower final linear-probing performance. With the RST-based curriculum, however, SiamJEPA largely closes this gap. Specifically, training with RST for the first 200 epochs and then continuing with the standard non-shuffled SiamJEPA objective for another 250 epochs achieves 73.3\% EMA/teacher linear-probing accuracy after 450 epochs in total. (Table~\ref{tab:position-decodability}). This result is comparable to the 72.9\% reported by I-JEPA and the 73.5\% reported by DSeq-JEPA, both after 600 training epochs, while requiring substantially fewer training epochs. Notably, unlike I-JEPA and DSeq-JEPA, SiamJEPA relies on a much simpler masking strategy, suggesting that the semantic-to-spatial curriculum can provide an effective alternative to more sophisticated masking or prediction schemes. Since Table~\ref{tab:performance_all}'s 12th-layer mean-pooling protocol differs from I-JEPA's own evaluation, we additionally re-evaluate the same checkpoint using I-JEPA's official linear-probing protocol (avg-pooled/4-layer-concat features, best of a 12-head grid): SiamJEPA reaches 74.2\% top-1, exceeding both I-JEPA (72.9\%) and DSeq-JEPA (73.5\%) even under a matched evaluation protocol, again with substantially fewer training epochs (450 vs.\ 600).

\begin{figure}[t]
  \centering
  \begin{subfigure}[t]{0.32\linewidth}
    \centering
    \includegraphics[width=.99\linewidth]{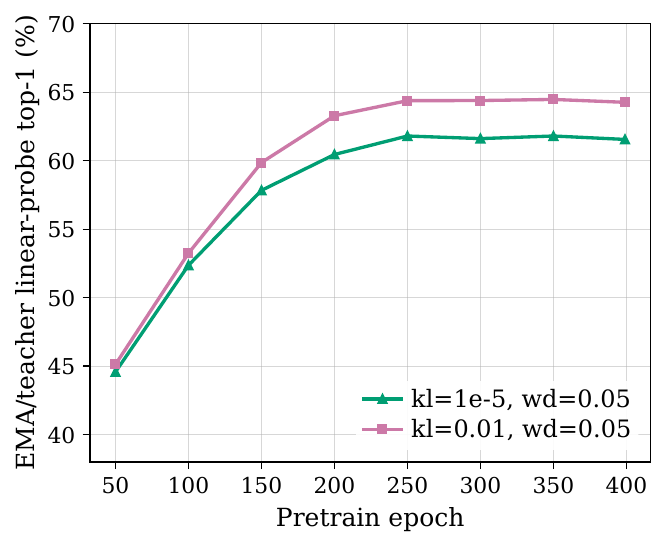}
    \caption{wd=0.05, cold-start}
    \label{subfig:kl-mask-a}
  \end{subfigure}
  \begin{subfigure}[t]{0.32\linewidth}
    \centering
    \includegraphics[width=.99\linewidth]{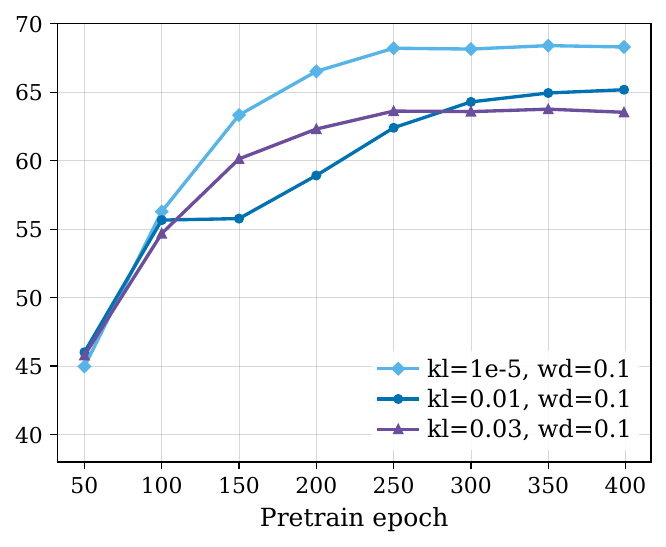}
    \caption{wd=0.1, cold-start}
    \label{subfig:kl-mask-b}
  \end{subfigure}
  \begin{subfigure}[t]{0.32\linewidth}
    \centering
    \includegraphics[width=.99\linewidth]{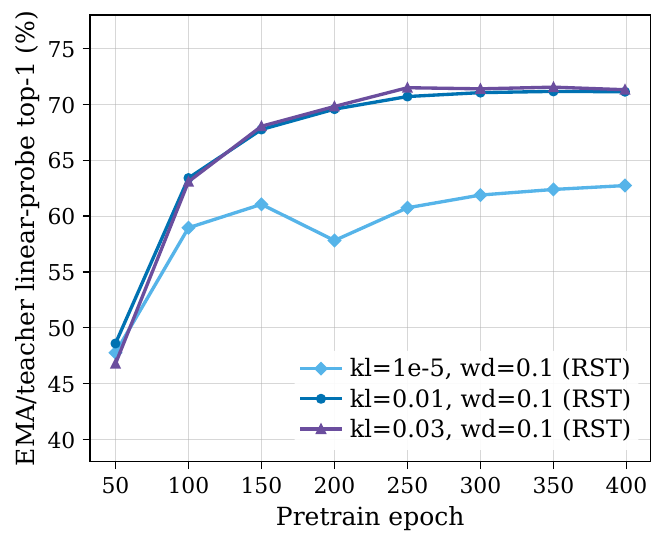}
    \caption{wd=0.1, RST regime}
    \label{subfig:kl-mask-e}
  \end{subfigure}

  \begin{subfigure}[t]{0.32\linewidth}
    \centering
    \includegraphics[width=.99\linewidth]{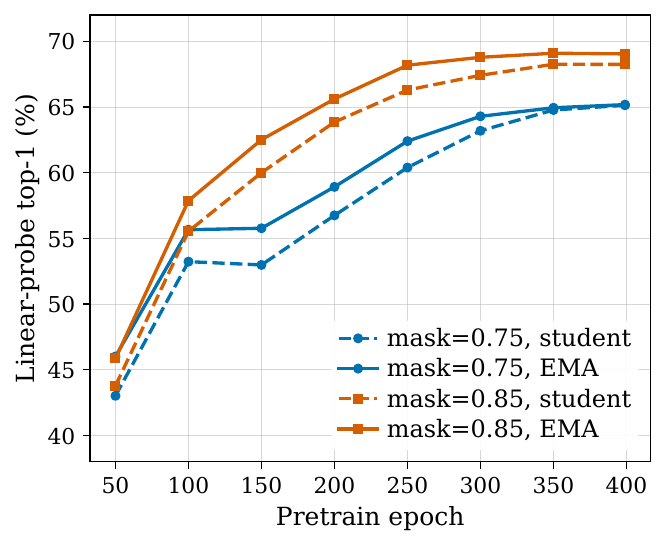}
    \caption{mask\_ratio (kl=0.01, wd=0.1)}
    \label{subfig:kl-mask-c}
  \end{subfigure}
  \begin{subfigure}[t]{0.32\linewidth}
    \centering
    \includegraphics[width=.99\linewidth]{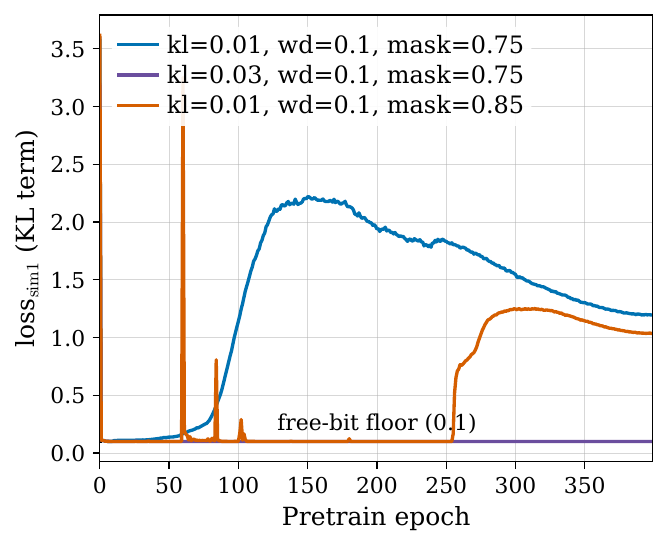}
    \caption{Training loss (loss$_{\mathrm{sim1}}$)}
    \label{subfig:kl-mask-d}
  \end{subfigure}
  \begin{subfigure}[t]{0.32\linewidth}
    \centering
    \includegraphics[width=.99\linewidth]{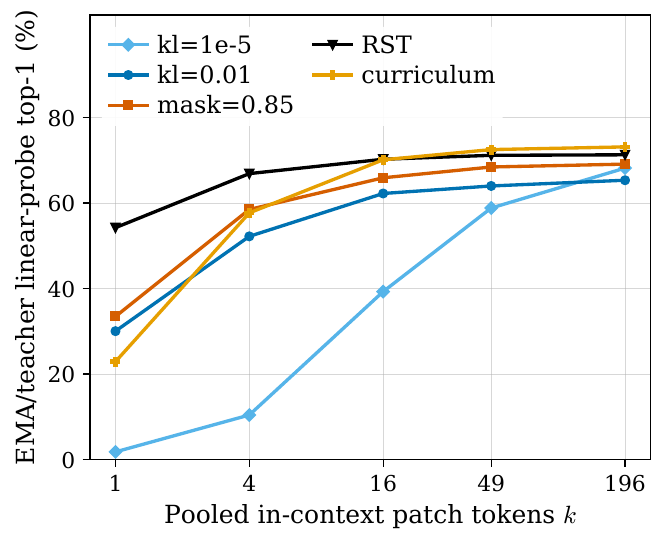}
    \caption{Class information per token}
    \label{subfig:kl-mask-f}
  \end{subfigure}
  \caption{Effect of the weight decay and mask ratio
    (cold-start unless noted). (a)--(d): EMA/teacher linear-probe top-1 vs.\
    pretrain epoch; (e): the corresponding loss$_{\mathrm{sim1}}$ (KL term)
    training dynamics for the configs in (b) and (d), bounded below by the
    free-bit floor of 0.1; (f): top-1 when only $k$ random patch tokens of
    the clean image are mean-pooled after the full encoder (EMA/teacher,
    wd=0.1, $\lambda_{\mathrm{KL}}=0.01$ and mask\_ratio=0.75 unless noted;
    RST at epoch 399; curriculum = init$\leftarrow$RST ep200 at 450 epochs;
    Appendix~\ref{sec:per-token-semantics}). (b) includes an
    intermediate free-bit scale ($\lambda_{\mathrm{KL}}=0.03$); (d) uses
    dashed=student, solid=EMA/teacher.}
  \label{fig:kl-wd}
  \label{fig:mask-analysis}
  \label{fig:rst-kl-wd01}
  \vspace{-.15in}
\end{figure}

\subsection{Effect of Siamese student encoder}
To investigate the role of the Siamese student encoders, we conduct an ablation study on the KL regularization weight $\lambda_{\mathrm{KL}}$, weight decay, and masking ratio. Figure~\ref{fig:kl-wd} summarizes the results without RST (mask ratio $=0.75$; panels (a)--(b)). With the default weight decay of $0.05$, increasing $\lambda_{\mathrm{KL}}$ from $10^{-5}$ to $0.01$ consistently improves linear-probing accuracy. Interestingly, this trend reverses with stronger weight decay ($\mathrm{wd}=0.1$): at epoch 399, $\lambda_{\mathrm{KL}}=10^{-5}$ outperforms $\lambda_{\mathrm{KL}}=0.01$. Panel (b) shows that linear-probing accuracy continues to improve under weak regularization, reaching $68.3\%$ by epoch 399, even as panel (e) shows its KL loss steadily increasing, suggesting that stronger student--student agreement is not always beneficial.

In contrast, strong KL regularization substantially improves performance when combined with RST (panel (c)), suggesting that Siamese consistency becomes particularly important under RST. Panel (d) further shows that increasing the mask ratio from $0.75$ to $0.85$ avoids the same saturation behavior, while the KL loss remains close to the free-bit floor, consistent with a smaller discrepancy between the two student views.

Overall, the benefit of Siamese consistency depends strongly on the training regime. Strong KL regularization is particularly effective with RST, but can be unnecessary or even detrimental when sufficient consistency is already maintained. These results suggest that the Siamese objective serves as a complementary consistency constraint rather than simply maximizing agreement between the student encoders.

\subsection{Random shuffle teacher experiments}
\label{sec:rst-method-experiments-body}

As shown in Figure~\ref{fig:rst-kl-wd01}(a)--(b), SiamJEPA's linear-probing performance without curriculum learning falls short of state-of-the-art methods, possibly because final-layer patch representations retain too much spatial and local information at the expense of semantics. Under RST (panel (e)), despite removing the teacher's spatial correspondence entirely, linear-probing performance is comparable to or better than the non-shuffled counterpart, suggesting that exact spatial correspondence is not necessary and that RST instead encourages more semantic patch representations.

\begin{figure}[t]
  \centering
  \begin{subfigure}[t]{0.48\linewidth}
    \centering
    \includegraphics[width=.99\linewidth]{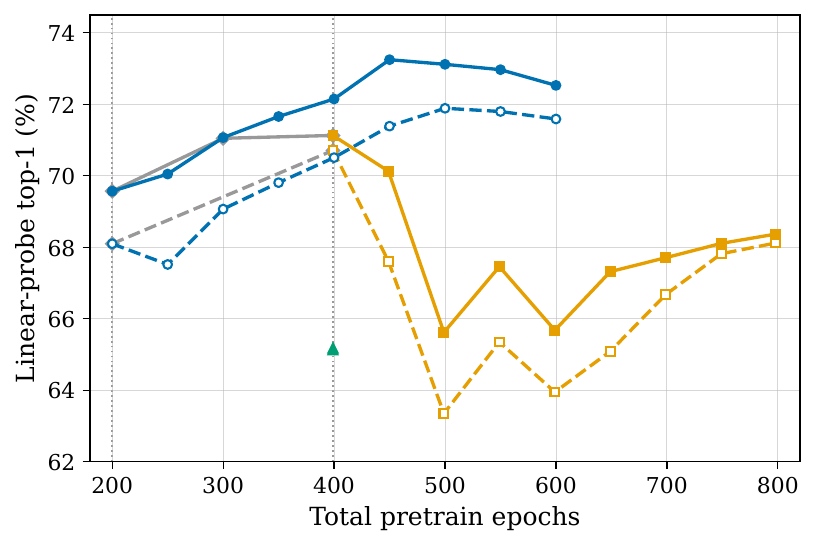}
    \caption{ImageNet linear probe}
    \label{subfig:position-curriculum-b}
  \end{subfigure}
  \begin{subfigure}[t]{0.48\linewidth}
    \centering
    \includegraphics[width=.99\linewidth]{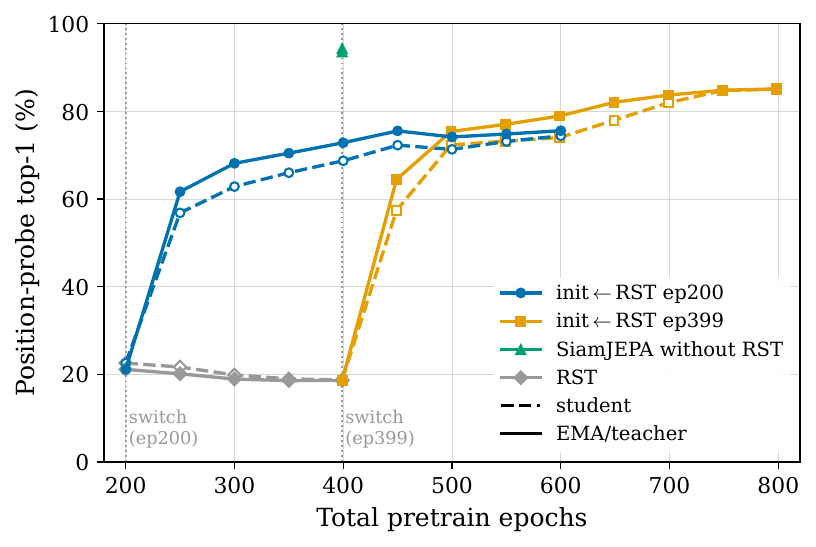}
    \caption{Position decodability}
    \label{subfig:position-curriculum-a}
  \end{subfigure}
  \caption{Position decodability and ImageNet linear-probe top-1 of the
    same checkpoints along the RST-based curriculum. Dotted vertical lines
    mark the switch from RST to the SiamJEPA-without-RST objective (epoch
    200 or 399). Dashed = student, solid = EMA/teacher. The position probe
    uses 2{,}000/500 images and a single seed (roughly $\pm1$ point).}
  \label{fig:position-curriculum}
  \vspace{-.2in}
\end{figure}

\begin{figure}[t]
  \centering
  \begin{subfigure}[t]{0.32\linewidth}
    \centering
    \includegraphics[width=.99\linewidth]{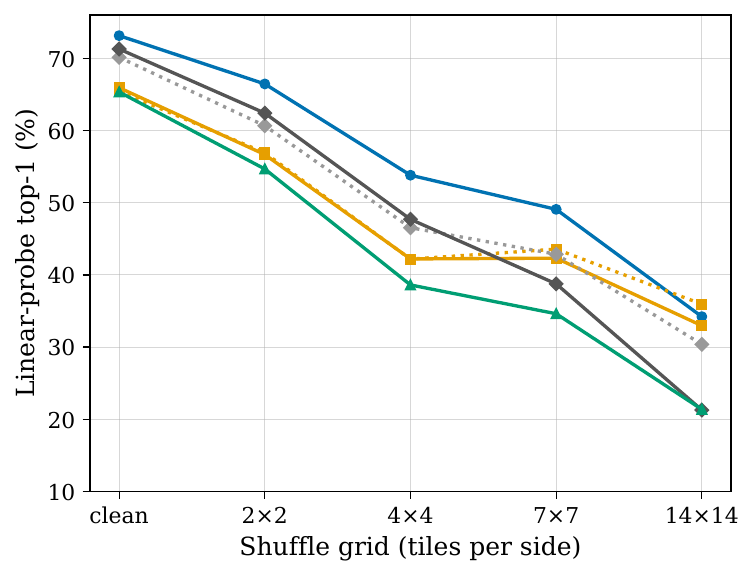}
    \caption{Patch shuffle}
    \label{subfig:ttr-a}
  \end{subfigure}
  \begin{subfigure}[t]{0.32\linewidth}
    \centering
    \includegraphics[width=.99\linewidth]{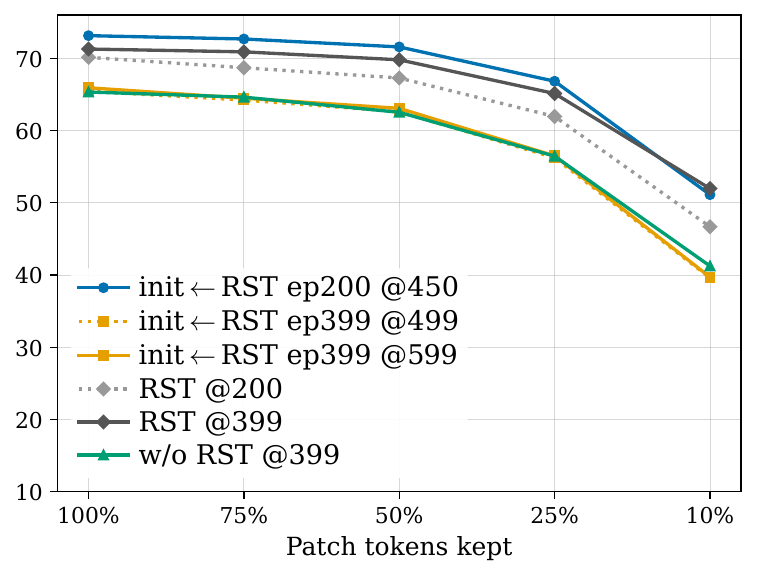}
    \caption{Token drop (random)}
    \label{subfig:ttr-b}
  \end{subfigure}
  \begin{subfigure}[t]{0.32\linewidth}
    \centering
    \includegraphics[width=.99\linewidth]{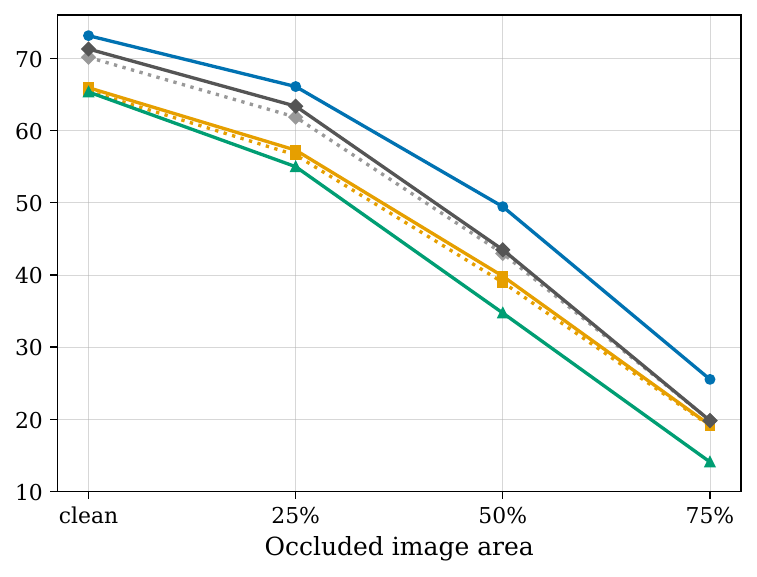}
    \caption{Occlusion (mean-colour block)}
    \label{subfig:ttr-c}
  \end{subfigure}
  \caption{Test-time robustness of the trained linear probes (EMA/teacher
    encoders, 10{,}000 ImageNet val images, no retraining; every random
    choice is shared across encoders). Patch shuffle permutes tiles of a
    $g\times g$ grid ($14\times14$ is a pure patch-token permutation);
    token drop feeds only a random fraction of patch tokens to the
    encoder; occlusion fills a square block of the given image area at a
    random location with the mean colour.}
  \label{fig:test-time-robustness}
  \vspace{-.15in}
\end{figure}

We verify this semantic bias directly with a linear probe that predicts a frozen final-layer patch token's original grid position out of $196$ classes (Appendix~\ref{sec:rst-ablation-appendix}, Table~\ref{tab:position-decodability}): SiamJEPA without RST retains almost perfect positional information ($94.3\%$, chance $0.51\%$), whereas RST loses most of it ($18.7\%$, still $37\times$ above chance), consistent with a shift toward DINO/BYOL-style global-content encoding. Figure~\ref{fig:position-curriculum} tracks this along the RST-based curriculum: positional decodability (b) stays low throughout RST training and recovers substantially once the objective switches back, at either switching point (epoch 200 or 399), while the corresponding ImageNet linear-probe accuracy (a) continues to improve. The same bias appears in test-time robustness (Figure~\ref{fig:test-time-robustness}): with only $10\%$ of patch tokens retained, RST retains $52.0\%$ top-1 versus $41.3\%$ without RST, showing RST relies less on specific spatial locations, though it is not completely position-agnostic.

Under $50\%$ image occlusion, the curriculum model (switched at epoch 200) is most robust ($49.5\%$), ahead of RST-only ($43.5\%$) and SiamJEPA without RST ($34.7\%$), indicating that the curriculum combines RST's robustness with the spatial information recovered afterward.

This bias is even visible at the single-token level (Figure~\ref{fig:kl-wd}(f)): a single random patch token already yields $54.2\%$ top-1 under RST, versus $30.0\%$ without RST, while a weak KL term ($\lambda_{\mathrm{KL}}=10^{-5}$) makes a single token almost uninformative ($1.8\%$, despite $68.2\%$ once averaged over all 196 tokens) -- indicating that RST and strong Siamese consistency independently push class information down into individual tokens. Further results are in Appendix~\ref{sec:rst-ablation-appendix}.

\subsection{Random shuffle curriculum learning}

One of our key findings is that RST rapidly improves linear probing performance during the early stages of training. However, its improvement tends to saturate in later epochs. In our experiments, using a larger weight decay alleviates this saturation and improves performance under longer training, but an additional mechanism appears necessary to further improve SiamJEPA. In contrast, the non-shuffled counterpart preserves spatial correspondence between teacher and student tokens, which may be beneficial for further refining the learned representations.

Figure~\ref{fig:position-curriculum} (a) reports the results of this curriculum learning strategy. As can be seen, the subsequent non-shuffled training continues to improve the representation rather than exhibiting the saturation observed with RST alone. These results suggest that semantic representation learning induced by RST and spatially aligned representation learning induced by the standard JEPA objective play complementary roles at different stages of training. As Figure~\ref{fig:position-curriculum} (b) above shows, positional information is lost under RST and recovered after switching back, but matching curricula on total training budget alone does not equalize final ImageNet accuracy, indicating that \emph{when} RST is switched off matters beyond how long it runs.

Interestingly, while the original SiamJEPA without curriculum learning underperforms I-JEPA and Dseq-JEPA with ViT-B, the proposed curriculum enables SiamJEPA to outperform I-JEPA in linear probing. Although our results are based on a single run and may therefore be affected by experimental randomness, they suggest that SiamJEPA can match or potentially outperform strong JEPA baselines while using a substantially simpler masking strategy.

\section{Conclusion}
In this paper, we investigated the role of Siamese student encoders within the Joint Embedding Predictive Architecture (JEPA) framework. To this end, we proposed SiamJEPA, a JEPA-based representation learning framework that incorporates Siamese student encoders together with an exponential moving average (EMA) teacher network. SiamJEPA processes two independently masked views using Siamese student encoders and learns to predict latent representations of masked target regions. Through extensive ablation studies on ImageNet linear probing with a ViT-Base backbone, we systematically investigated the effects of key design choices, including KL regularization, masking strategies, weight decay, and the free-bit threshold. Our results demonstrate that incorporating Siamese student encoders consistently improves representation quality over a corresponding JEPA-like baseline. Moreover, we find that the Siamese architecture serves as an effective regularizer, leading to more discriminative latent representations and faster convergence during the early stages of training.

Overall, our findings provide new insights into the role of Siamese student encoders in JEPA and offer practical guidance for designing and optimizing future JEPA-style representation learning methods.

\section{Future work}
Several important directions remain for future work. First, our study primarily focused on relatively small-scale architectures to better understand the role of Siamese student encoders. Whether the observed benefits extend to larger-scale models, such as ViT-Large and ViT-Huge, remains an important open question.

Second, we found that the final performance is highly sensitive to implementation details and hyperparameter choices. Although our experimental setup consistently outperformed masked autoencoder baselines, there is likely substantial room for further improvement through architectural refinements and more systematic hyperparameter optimization. 

Third, this work focused exclusively on image-based pretraining. Previous studies have shown that Siamese student encoders are particularly effective for video representation learning \citep{gupta2023siamese,jang2024visual,yamada2025phinet}. Moreover, as JEPA-based approaches have rapidly gained attention, several recent works have begun incorporating Siamese student encoder designs into video predictive learning frameworks \citep{daithankar2026you,rao2026skyjepa}. These developments suggest that a deeper understanding of Siamese student encoders may have broader implications beyond image representation learning, particularly for video understanding and world modeling.

Finally, SiamJEPA is inspired by the brain-inspired representation learning framework PhiNets \citep{ishikawa2025phinets,yamada2025phinet}. Exploring additional neuroscience-inspired mechanisms may further improve representation quality while providing new insights into the relationship between biological and artificial learning systems.

\subsection*{AI use statement}
We used ChatGPT and Claude Code to assist with implementation and debugging, experiments including hyperparameter tuning, literature searches, manuscript drafting and revision, and figure preparation. The authors take full responsibility for the final manuscript, code, references, experimental results, and reported findings.

\bibliography{ref}

@article{he2025dseq,
  title = {{DSeq-JEPA}: Discriminative Sequential {Joint-Embedding} Predictive Architecture},
  author = {Xiangteng He and Shunsuke Sakai and Chandhok, Shivam and Beery, Sara and Kun Yuan and Nicolas Padoy and Tatsuhito Hasegawa and Leonid Sigal},
  journal = {ECCV},
  year = {2026}
}

@article{wu2026visreg,
  title = {{VISReg}: {Variance-Invariance-Sketching} Regularization for {JEPA} training},
  author = {Haiyu Wu and Randall Balestriero and Morgan Levine},
  journal = {arXiv},
  year = {2026},
  doi = {10.48550/arXiv.2606.02572},
  url = {https://arxiv.org/pdf/2606.02572.pdf},
}

@article{assran2025v,
  title={V-jepa 2: Self-supervised video models enable understanding, prediction and planning},
  author={Assran, Mido and Bardes, Adrien and Fan, David and Garrido, Quentin and Howes, Russell and Muckley, Matthew and Rizvi, Ammar and Roberts, Claire and Sinha, Koustuv and Zholus, Artem and others},
  journal={arXiv preprint arXiv:2506.09985},
  year={2025}
}

@article{bardes2024revisiting,
  title={Revisiting feature prediction for learning visual representations from video},
  author={Bardes, Adrien and Garrido, Quentin and Ponce, Jean and Chen, Xinlei and Rabbat, Michael and LeCun, Yann and Assran, Mahmoud and Ballas, Nicolas},
  journal={arXiv preprint arXiv:2404.08471},
  year={2024}
}

@inproceedings{assran2023self,
  title={Self-supervised learning from images with a joint-embedding predictive architecture},
  author={Assran, Mahmoud and Duval, Quentin and Misra, Ishan and Bojanowski, Piotr and Vincent, Pascal and Rabbat, Michael and LeCun, Yann and Ballas, Nicolas},
  booktitle={CVPR},
  year={2023}
}

@article{simeoni2025dinov3,
  title={Dinov3},
  author={Sim{\'e}oni, Oriane and Vo, Huy V and Seitzer, Maximilian and Baldassarre, Federico and Oquab, Maxime and Jose, Cijo and Khalidov, Vasil and Szafraniec, Marc and Yi, Seungeun and Ramamonjisoa, Micha{\"e}l and others},
  journal={arXiv preprint arXiv:2508.10104},
  year={2025}
}

@article{becker1992self,
  title     = {Self-organizing neural network that discovers surfaces in random-dot stereograms},
  author    = {Becker, Suzanna and Hinton, Geoffrey E},
  journal   = {Nature},
  volume    = {355},
  number    = {6356},
  pages     = {161--163},
  year      = {1992},
  publisher = {Nature Publishing Group UK London}
}

@inproceedings{chen2020simple,
  title     = {A simple framework for contrastive learning of visual representations},
  author    = {Chen, Ting and Kornblith, Simon and Norouzi, Mohammad and Hinton, Geoffrey},
  booktitle = {ICML},
  year      = {2020}
}

@article{vaswani2017attention,
  title   = {Attention is all you need},
  author  = {Vaswani, Ashish and Shazeer, Noam and Parmar, Niki and Uszkoreit, Jakob and Jones, Llion and Gomez, Aidan N and Kaiser, {\L}ukasz and Polosukhin, Illia},
  journal = {NIPS},
  year    = {2017}
}

@inproceedings{chen2021exploring,
  title={Exploring simple {Siamese} representation learning},
  author={Chen, Xinlei and He, Kaiming},
  booktitle={CVPR},
  year={2021}
}

@inproceedings{caron2021emerging,
  title={Emerging properties in self-supervised vision transformers},
  author={Caron, Mathilde and Touvron, Hugo and Misra, Ishan and J{\'e}gou, Herv{\'e} and Mairal, Julien and Bojanowski, Piotr and Joulin, Armand},
  booktitle={ICCV},
  year={2021}
}

@inproceedings{grill2020bootstrap,
  title   = {Bootstrap your own latent-a new approach to self-supervised learning},
  author  = {Grill, Jean-Bastien and Strub, Florian and Altch\'{e}, Florent and Tallec, Corentin and Richemond, Pierre and Buchatskaya, Elena and Doersch, Carl and Avila Pires, Bernardo and Guo, Zhaohan and Gheshlaghi Azar, Mohammad and Piot, Bilal and kavukcuoglu, koray and Munos, Remi and Valko, Michal},
  booktitle = {NeurIPS},
  year    = {2020}
}

@article{chen2020improved,
  title   = {Improved baselines with momentum contrastive learning},
  author  = {Chen, Xinlei and Fan, Haoqi and Girshick, Ross and He, Kaiming},
  journal = {arXiv preprint arXiv:2003.04297},
  year    = {2020}
}

@article{yamada2025phinet,
  title={Brain-Inspired Stochastic Joint Embedding Representation Learning},
  author={Yamada, Makoto and Chai, Kian Ming A and Rhim, Ayoub and Ishikawa, Satoki and Sabokrou, Mohammad and Tsai, Yao-Hung Hubert},
  journal={arXiv preprint arXiv:2505.11129},
  year={2025}
}

@inproceedings{zbontar2021barlow,
  title     = {{Barlow Twins}: Self-supervised learning via redundancy reduction},
  author    = {Zbontar, Jure and Jing, Li and Misra, Ishan and LeCun, Yann and Deny, St{\'e}phane},
  booktitle = {ICML},
  year      = {2021}
}

@article{chen2022predictive,
  title={Predictive Sequence Learning in the Hippocampal Formation},
  author={Chen, Yusi and Zhang, Huanqiu and Cameron, Mia and Sejnowski, Terrrence},
  volume={112},
  journal={Neuron},
  pages={2645–2658},
  year={2024},
}

@article{mcclelland1995there,
  title={Why there are complementary learning systems in the hippocampus and neocortex: insights from the successes and failures of connectionist models of learning and memory.},
  author={McClelland, James L and McNaughton, Bruce L and O'Reilly, Randall C},
  journal={Psychological Review},
  volume={102},
  number={3},
  pages={419},
  year={1995},
  publisher={American Psychological Association}
}

@inproceedings{
jang2024visual,
title={Visual Representation Learning with Stochastic Frame Prediction},
author={Huiwon Jang and Dongyoung Kim and Junsu Kim and Jinwoo Shin and Pieter Abbeel and Younggyo Seo},
booktitle={ICMLR},
year={2024}
}

@article{daithankar2026you,
  title={You Don't Need Strong Assumptions: Visual Representation Learning via Temporal Differences},
  author={Daithankar, Ninad and Gladstone, Alexi and LeCun, Yann and Ji, Heng},
  journal={arXiv preprint arXiv:2606.15956},
  year={2026}
}

@article{rao2026skyjepa,
  title={SkyJEPA: Learning Long-Horizon World Models for Zero-Shot Sim-to-Real Control of Quadrotors},
  author={Rao, Pratyaksh and Zhang, Wancong and Balestriero, Randall and LeCun, Yann and Loianno, Giuseppe},
  journal={arXiv preprint arXiv:2606.23444},
  year={2026}
}

@inproceedings{
ishikawa2025phinets,
title={{PhiNets}: Brain-inspired Non-contrastive Learning Based on Temporal Prediction Hypothesis},
author={Satoki Ishikawa and Makoto Yamada and Han Bao and Yuki Takezawa},
booktitle={ICLR},
year={2025}
}

@inproceedings{
dosovitskiy2021an,
title={An Image is Worth 16x16 Words: Transformers for Image Recognition at Scale},
author={Alexey Dosovitskiy and Lucas Beyer and Alexander Kolesnikov and Dirk Weissenborn and Xiaohua Zhai and Thomas Unterthiner and Mostafa Dehghani and Matthias Minderer and Georg Heigold and Sylvain Gelly and Jakob Uszkoreit and Neil Houlsby},
booktitle={ICLR},
year={2021}
}

@inproceedings{cropmae,
author = {Eyma\"{e}l, Alexandre and Vandeghen, Renaud and Cioppa, Anthony and Giancola, Silvio and Ghanem, Bernard and Van Droogenbroeck, Marc},
title = {Efficient Image Pre-training with {Siamese} Cropped Masked Autoencoders},
year = {2024},
booktitle = {ECCV}
}

@inproceedings{hafner2020dreamerv2,
  title={Mastering {A}tari with Discrete World Models},
  author={Hafner, Danijar and Lillicrap, Timothy and Norouzi, Mohammad and Ba, Jimmy},
 booktitle={ICLR},
  year={2021}
}

@InProceedings{pmlr-v80-denton18a,
  title = 	 {Stochastic Video Generation with a Learned Prior},
  author =       {Denton, Emily and Fergus, Rob},
  booktitle = 	 {ICML},
  year = 	 {2018}
}

@article{gupta2023siamese,
  title={Siamese masked autoencoders},
  author={Gupta, Agrim and Wu, Jiajun and Deng, Jia and Li, Fei-Fei},
  journal={NeurIPS},
  year={2023}
}

@article{balestriero2025lejepa,
  title={Le{JEPA}: Provable and scalable self-supervised learning without the heuristics},
  author={Balestriero, Randall and LeCun, Yann},
  journal={arXiv preprint arXiv:2511.08544},
  year={2025}
}

@article{zemel1990discovering,
  title={Discovering viewpoint-invariant relationships that characterize objects},
  author={Zemel, Richard and Hinton, Geoffrey E},
  journal={NIPS},
  year={1990}
}

@article{lecun2022path,
  title={A path towards autonomous machine intelligence version 0.9. 2, 2022-06-27},
  author={LeCun, Yann and others},
  journal={Open Review},
  volume={62},
  number={1},
  pages={1--62},
  year={2022}
}

@article{schmidhuber1993discovering,
  title={Discovering predictable classifications},
  author={Schmidhuber, J{\"u}rgen and Prelinger, Daniel},
  journal={Neural Computation},
  volume={5},
  number={4},
  pages={625--635},
  year={1993},
  publisher={MIT Press}
}

@article{chen2024context,
  title={Context autoencoder for self-supervised representation learning},
  author={Chen, Xiaokang and Ding, Mingyu and Wang, Xiaodi and Xin, Ying and Mo, Shentong and Wang, Yunhao and Han, Shumin and Luo, Ping and Zeng, Gang and Wang, Jingdong},
  journal={International Journal of Computer Vision},
  volume={132},
  number={1},
  pages={208--223},
  year={2024}
}

@inproceedings{assran2022masked,
  title={Masked siamese networks for label-efficient learning},
  author={Assran, Mahmoud and Caron, Mathilde and Misra, Ishan and Bojanowski, Piotr and Bordes, Florian and Vincent, Pascal and Joulin, Armand and Rabbat, Mike and Ballas, Nicolas},
  booktitle={ECCV},
  year={2022}
}

@inproceedings{he2022masked,
  title={Masked autoencoders are scalable vision learners},
  author={He, Kaiming and Chen, Xinlei and Xie, Saining and Li, Yanghao and Doll{\'a}r, Piotr and Girshick, Ross},
  booktitle={CVPR},
  year={2022}
}

@inproceedings{
bardes2022vicreg,
title={{VICR}eg: Variance-Invariance-Covariance Regularization for Self-Supervised Learning},
author={Adrien Bardes and Jean Ponce and Yann LeCun},
booktitle={ICLR},
year={2022},
xurl={https://openreview.net/forum?id=xm6YD62D1Ub}
}

@article{
oquab2024dinov,
title={{DINO}v2: Learning Robust Visual Features without Supervision},
author={Maxime Oquab and Timoth{\'e}e Darcet and Th{\'e}o Moutakanni and Huy V. Vo and Marc Szafraniec and Vasil Khalidov and Pierre Fernandez and Daniel HAZIZA and Francisco Massa and Alaaeldin El-Nouby and Mido Assran and Nicolas Ballas and Wojciech Galuba and Russell Howes and Po-Yao Huang and Shang-Wen Li and Ishan Misra and Michael Rabbat and Vasu Sharma and Gabriel Synnaeve and Hu Xu and Herve Jegou and Julien Mairal and Patrick Labatut and Armand Joulin and Piotr Bojanowski},
journal={Transactions on Machine Learning Research},
year={2024}
}

\clearpage
\appendix

\begin{table}[t]
\centering
\caption{Pretraining and linear probing configurations.}
\label{tab:config}
\begin{tabular}{lcc}
\hline
\textbf{Setting} & \textbf{MAE} & \textbf{SiamJEPA} \\
\hline
Backbone & ViT-Base & ViT-Base \\
Decoder depth & 8 & 1\\
Pretrain Dataset & ImageNet-1K & ImageNet-1K \\
Pretrain Epochs & 400 & 400 \\
Input Resolution & $224^2$ & $224^2$ \\
Mask Ratio & 0.75 & \{0.75,0.85\} \\
Optimizer & AdamW & AdamW \\
Base Learning Rate & 1.5e-4 & 1.5e-4 \\
Weight Decay & 0.05 & \{0.05,0.1\} \\
Effective Batch Size & 4096 & 8192 \\
EMA & No & \{0.99, 0.999, 0.9999\} \\
\hline
Linear Probe Epochs & 90 & 90 \\
Linear Probe Resolution & $224^2$ & $224^2$ \\
\hline
\end{tabular}
\end{table}

\section{Extended related work}

\noindent {\bf Joint embedding architecture:} One of the most popular SSL approaches is contrastive learning. SimCLR \citep{chen2020simple} learns representations by pulling together augmented views of the same image (positive pairs) while pushing apart different images (negative pairs), but relies on large negative-sample counts and thus benefits from very large batch sizes. Momentum Contrast (MoCo) \citep{chen2020improved} alleviates this with a momentum encoder and a fixed-size negative-sample queue.

Subsequent work showed negative samples are not strictly necessary: BYOL \citep{grill2020bootstrap} and SimSiam \citep{chen2021exploring} achieve strong performance via stop-gradient operations and exponential moving averages instead. Another family relies on explicit regularization, e.g., Barlow Twins \citep{zbontar2021barlow} and Variance-Invariance-Covariance Regularization (VICReg) \citep{bardes2022vicreg}, which encourage informative, non-collapsed representations through variance and covariance constraints. Masked Siamese Networks (MSN) \citep{assran2022masked} extend BYOL to masked modeling, using a student and an EMA teacher encoder that align CLS-token embeddings, and outperform BYOL particularly in the 1\% ImageNet-1K label-efficient setting.

PhiNet \citep{ishikawa2025phinets}, a brain-inspired SSL framework drawing on the biological circuitry of the hippocampus and neocortex, employs Siamese encoders with temporal prediction and EMA mechanisms, and is more robust to collapse than SimSiam with favorable continual-learning properties. Unlike other methods, which use a single student encoder, PhiNet uses Siamese student encoders. 

\noindent {\bf Masked modeling with pixel reconstruction:} Most of the aforementioned methods were originally built on CNNs such as ResNet; Transformer-based SSL has since become dominant. A seminal example is Masked Autoencoders (MAE) \citep{he2022masked}, which reconstruct masked image patches from visible tokens. Another line combines masked modeling with Siamese architectures: SiamMAE \citep{gupta2023siamese} predicts future video frames from the current frame and a masked future observation via two encoders and a decoder, and CropMAE \citep{cropmae} and RSP \citep{jang2024visual} follow similarly, showing strong performance on downstream tasks such as pose estimation and segmentation. However, these methods remain fundamentally reconstruction-based, aiming to recover image content.

\section{Ablation study}

\subsection{Effect of weight decay}
Next, we investigate the effect of weight decay by varying its coefficient, as shown in Figure~\ref{fig:kl-wd}. Consistent with the observations reported for PhiNet \citep{ishikawa2025phinets}, SiamJEPA is relatively robust to the choice of weight decay. At $\lambda_{\mathrm{KL}}=0.01$, larger weight decay is particularly beneficial during longer training: with the default wd=0.05, linear probing performance saturates by epoch 300 (63.56\%$\to$63.78\%$\to$63.59\% at epochs 300/350/399), whereas wd=0.1 continues improving through epoch 399 (63.19\%$\to$64.77\%$\to$65.13\%), eventually overtaking wd=0.05. This benefit is not restricted to strong KL regularization: at $\lambda_{\mathrm{KL}}=10^{-5}$, wd=0.1 is, if anything, even more advantageous, reaching 67.53\% at epoch 399 (fully converged) versus wd=0.05's final 60.84\%. This is notably different from an earlier, pre-fix observation of training instability for $\lambda_{\mathrm{KL}}=10^{-5}$ under wd=0.1 (Appendix Section~\ref{sec:rst-ablation-appendix}); after the \texttt{forward\_encoder} mask-shuffle fix, we no longer observe this failure mode, and $\lambda_{\mathrm{KL}}=10^{-5}$/wd=0.1 is in fact the strongest cold-start cell in the grid so far.

These results suggest that stronger weight decay improves generalization during extended pre-training for SiamJEPA, and that this benefit holds across KL regularization strengths rather than being specific to strongly KL-regularized configurations.

\subsection{Per-token semantic information}
\label{sec:per-token-semantics}
To measure how much class information individual patch tokens carry, we pass the clean image through the frozen encoder and mean-pool only $k$ randomly chosen patch tokens before applying the trained linear probe (Figure~\ref{fig:kl-wd}(f); EMA/teacher encoders, 10{,}000 ImageNet val images, the same random tokens for all encoders). Unlike token dropping, the attention context of all tokens is kept intact. Because the probe was trained on 196-token means, small $k$ is out of distribution for the probe, so the numbers are best read as a ranking across encoders. RST stands out: a single patch token already yields $54.2\%$ top-1 accuracy ($76\%$ of its clean accuracy), compared with $30.0\%$ for SiamJEPA without RST, directly supporting the view that RST spreads global semantic information over individual patch tokens. A higher mask ratio (0.85) barely changes the per-token information ($33.5\%$), so its robustness to occlusion has a different origin. The curriculum model reads less class information from a single token ($22.9\%$), presumably because positional information is re-introduced into each token in the second stage, yet it matches RST with 16 tokens ($70.1\%$). With a weak KL term ($\lambda_{\mathrm{KL}}=10^{-5}$), single tokens are almost uninformative to the probe ($1.8\%$) although the full average reaches $68.2\%$; this encoder is also the most dependent on the position embedding, which suggests large token-specific positional components that cancel only in the average.

\section{Random shuffle teacher ablation study}
\label{sec:rst-ablation-appendix}

The ablations reported in this section were all conducted under the Random Shuffle Teacher (RST) training regime introduced in Section~\ref{sec:rst-method}, in which the teacher's patch tokens are randomly permuted before the prediction objective is computed. These sweeps additionally use mean pooling from an intermediate (10th) transformer layer as their evaluation protocol.

\subsection{Position decodability and semantic bias of RST}

We test whether shuffling the teacher's patch order washes out positional information by training a linear probe on frozen final-layer patch tokens (pre-pool, no \texttt{[CLS]}) to predict the token's original grid position out of $196$ classes. Table~\ref{tab:position-decodability} compares checkpoints trained under identical hyperparameters ($\mathrm{KL}=0.01$, $\mathrm{wd}=0.1$), differing only in whether \texttt{forward\_encoder} shuffled patch order for the teacher branch (RST) or preserved it (SiamJEPA without RST), together with two curriculum models initialized from RST at epoch 200 and 399. SiamJEPA without RST retains almost perfect positional information ($94.3\%$ top-1, chance $0.51\%$), while RST loses most of it ($18.7\%$ top-1) though it remains roughly $37\times$ above chance, indicating the representation is not fully scrambled. This supports the hypothesis that shuffling the teacher's patch order during pretraining biases the learned representation toward a more DINO/BYOL-style global-content encoding at the expense of spatial position.

\begin{table}[t]
\centering
\small
\caption{Position-decodability of frozen final-layer patch tokens. A linear
probe predicts which of the $196$ spatial grid positions a frozen patch token
was extracted from (chance $1/196 \approx 0.51\%$). All encoders use
$\mathrm{KL}=0.01$, $\mathrm{wd}=0.1$, mask\_ratio$=0.75$, and differ only in
whether \texttt{forward\_encoder} shuffled the teacher's patch order during
pretraining, and whether the run was subsequently continued under the
SiamJEPA-without-RST (non-shuffled) objective. The last column repeats the ImageNet
linear-probe top-1 of the very same checkpoint, so that positional and
semantic quality can be read side by side.}
\label{tab:position-decodability}
\begin{tabular}{llccc c}
\toprule
Encoder & Branch & Epoch & Pos.\ top-1 (\%) & Pos.\ top-5 (\%) & IN-1k  top-1 (\%) \\
\midrule
SiamJEPA without RST  & student     & 399 & \textbf{94.3} & \textbf{99.9} & 65.13 \\
                          & EMA/teacher & 399 & 93.5          & 99.9          & 65.18 \\
\addlinespace[2pt]
RST & student     & 399 & 18.7 & 57.6 & 70.71 \\
                             & EMA/teacher & 399 & 18.6 & 57.2 & 71.13 \\
\addlinespace[2pt]
init$\leftarrow$RST ep200  & student     & 450 & 72.3 & 99.0 & 71.39 \\
                                       & EMA/teacher & 450 & 75.5 & 99.3 & {\bf 73.25} \\
\addlinespace[2pt]
init$\leftarrow$RST ep399  & student     & 599 & 74.0 & 99.1 & 63.95 \\
                                       & EMA/teacher & 599 & 78.9 & 99.3 & 65.68 \\
\bottomrule
\end{tabular}
\end{table}

\begin{table}[t]
\centering
\caption{Additional ways of leveraging the pre-fix RST (Random Shuffle Teacher) checkpoint, extending Table~\ref{tab:position-decodability} with configurations not otherwise tabulated there (SiamJEPA without RST, RST, and init$\leftarrow$RST ep200 @450 already appear in Table~\ref{tab:position-decodability}). init$\leftarrow$ = fresh optimizer/epoch-0 restart (\texttt{--init\_checkpoint}); resume$\leftarrow$ = optimizer/epoch-preserving continuation (\texttt{--resume}).}
\label{tab:rst}
\begin{tabular}{lccc}
\toprule
Configuration & Epoch & Student top-1 (\%) & EMA/teacher top-1 (\%) \\
\midrule
resume$\leftarrow$RST ep200 & 399 & 71.13 & 71.86 \\
init$\leftarrow$RST ep200 & 599 & 71.59 & 72.53 \\
init$\leftarrow$RST ep399 & 799 & 68.12 & 68.37 \\
\bottomrule
\end{tabular}
\end{table}

\begin{figure}[t]
  \centering
  \begin{subfigure}[t]{0.48\linewidth}
    \centering
    \includegraphics[width=.99\linewidth]{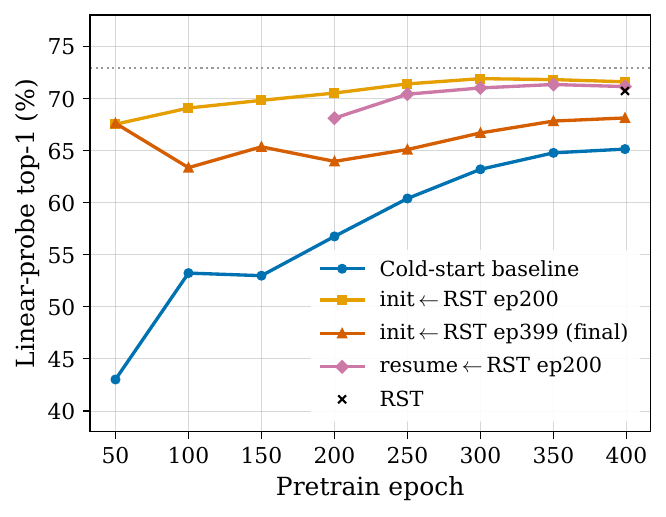}
    \caption{Student}
    \label{subfig:rst-student}
  \end{subfigure}
  \begin{subfigure}[t]{0.48\linewidth}
    \centering
    \includegraphics[width=.99\linewidth]{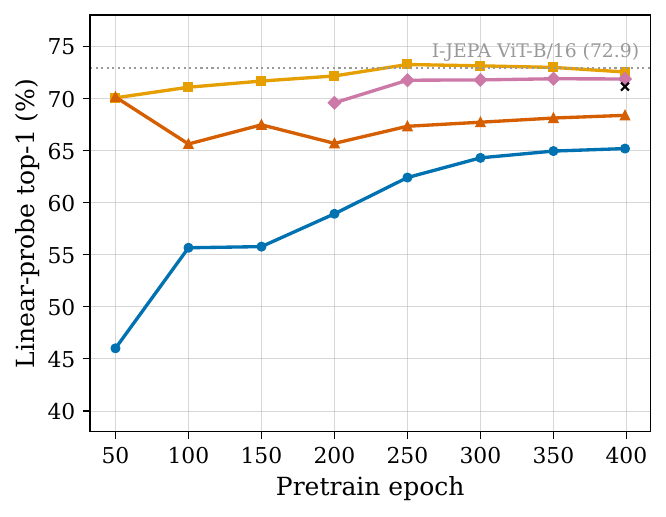}
    \caption{EMA / teacher}
    \label{subfig:rst-ema}
  \end{subfigure}
  \caption{Ways of leveraging the pre-fix RST (Random Shuffle Teacher)
    checkpoint after the encoder-mask-shuffle fix, compared against
    SiamJEPA without RST trained from scratch (cold-start) and the
    uncorrected RST run itself. Dotted line:
    I-JEPA ViT-B/16 reference (72.9\%).}
  \label{fig:rst}
  \vspace{-.15in}
\end{figure}

Figure~\ref{fig:rst} visualizes the additional configurations of Table~\ref{tab:rst} across the full pretrain trajectory (student and EMA/teacher), rather than only at their final checkpoints.

\begin{figure*}[t!]
    \centering
      \begin{subfigure}[t]{0.45\textwidth}
        \centering
        \includegraphics[width=.99\textwidth]{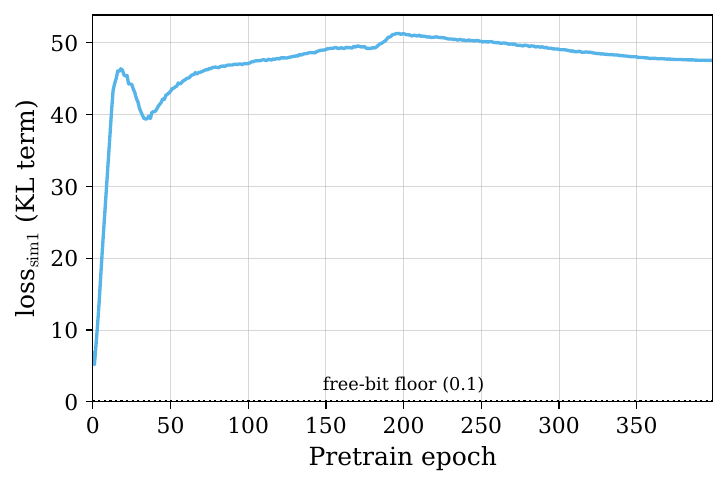}%
        \caption{$\lambda_{\textnormal{KL}}=0.00001$.\label{subfig:loss_00001}}
    \end{subfigure}
        \begin{subfigure}[t]{0.45\textwidth}
        \centering
        \includegraphics[width=.99\textwidth]{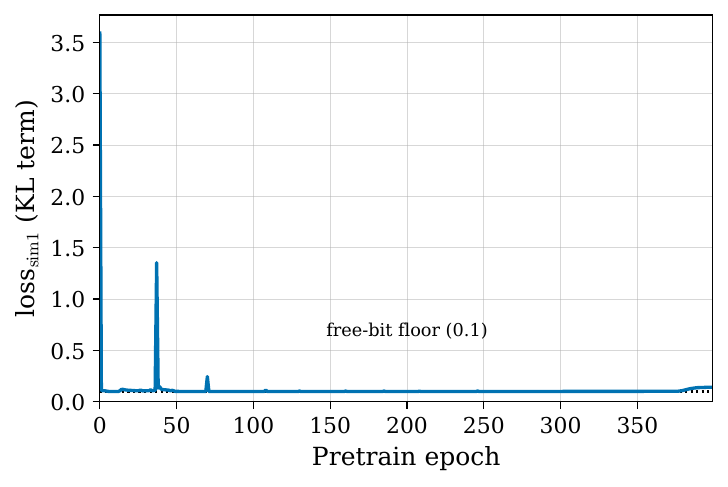}%
        \caption{$\lambda_{\textnormal{KL}}=0.01$. \label{subfig:loss_001}}
    \end{subfigure}
    \caption{RST-regime training-loss$_{\mathrm{sim1}}$ (KL term) dynamics for $\lambda_{\textnormal{KL}}=0.00001$ and $\lambda_{\textnormal{KL}}=0.01$ (both wd=0.1, mask\_ratio=0.75). With a small regularization coefficient, the KL divergence between the representations produced by the two Siamese student encoders remains large throughout training (rising above 40). In contrast, with a larger regularization coefficient, the KL divergence stays close to the free-bit threshold (0.1 in our experiments) for essentially the whole 399 epochs, aside from brief transient spikes.}
    \label{fig:learning_curve}
\end{figure*}

\subsection{Effect of Siamese student encoder for RST}
To investigate the role of the Siamese student encoders for RST, we conduct an ablation study on the KL regularization weight $\lambda_{\mathrm{KL}}$, the weight decay parameter,  and the free-bit threshold used in the KL divergence computation. The KL term encourages the outputs of the two Siamese student encoders to become similar, while the free-bit threshold prevents the two distributions from becoming completely identical by imposing a minimum KL value.

Figure \ref{fig:learning_curve} shows the learning curves for small and large values of $\lambda_{\textnormal{KL}}$. As expected, a small regularization weight results in a large KL divergence (greater than 40), whereas a large regularization weight keeps the KL divergence close to the free-bit threshold of 0.1. Despite this substantial difference in the KL term, the overall learning behavior of the total loss is similar in both cases, although training with the larger regularization weight converges more slowly. These observations suggest that the primary learning signal for acquiring useful representations comes from the masked prediction objective (i.e., the Sim-2 loss), whereas the KL term mainly serves as a regularizer. In particular, the KL regularization provides a beneficial inductive bias during the early stages of training, but overly strong regularization may unnecessarily constrain optimization and slow convergence.

\subsection{Predictor depth}
The effect of predictor depth remains an open question. Owing to computational constraints and the large hyperparameter space, we have not yet been able to determine whether the observations described below are intrinsic properties of SiamJEPA (RST) or are specific to our experimental setting.

Empirically, we found that SiamJEPA (RST) achieves better linear probing performance when using a shallow Transformer predictor with only one or two layers. Throughout this paper, we therefore employ a single-layer predictor, which is substantially shallower than those commonly used in JEPA-based methods. If this observation generalizes beyond our current setting, it would suggest that SiamJEPA (RST) can achieve competitive performance with a significantly smaller predictor, leading to a more parameter-efficient architecture.

We also observed that the final linear probing performance depends on which encoder layer is used for evaluation. This suggests that the layer containing the most transferable representation may vary with the predictor depth, consistent with the observations in the previous section. It is possible that this behavior changes for larger models or more challenging datasets. However, thoroughly investigating this phenomenon would require substantially more computational resources than were available in this work. We therefore leave a systematic study of predictor depth and representation dynamics for future work.

\subsection{Test-time robustness}

Figure~\ref{fig:test-time-robustness} evaluates the trained linear probes under test-time perturbations without any retraining. RST is the most robust to dropping patch tokens: with only $10\%$ of the tokens kept, it retains $52.0\%$ top-1 accuracy ($73\%$ of its clean accuracy), compared with $41.3\%$ ($63\%$) for SiamJEPA without RST, consistent with semantic information being spread across patch tokens. However, RST is not position-agnostic. Setting the patch position embedding to zero reduces its accuracy from $71.3\%$ to $16.4\%$, a larger drop than for SiamJEPA without RST ($65.3\%\to21.9\%$), and a $14\times14$ patch shuffle, which only reassigns position embeddings to patch contents, is as harmful for RST as for SiamJEPA without RST ($21.3\%$ vs.\ $21.4\%$). Thus, RST removes positional information from the final-layer representation, while its intermediate computation still relies on the position embedding, increasingly so with longer RST training ($34.6\%$ at epoch 200 vs.\ $16.4\%$ at epoch 399 without the position embedding). Under occlusion with a mean-colour block, which keeps all tokens and is closest to a real occluder, the curriculum model switched at epoch 200 is clearly the most robust ($49.5\%$ at $50\%$ occluded area, vs.\ $43.5\%$ for RST and $34.7\%$ for SiamJEPA without RST).

% Appendix: qualitative analysis of [CLS] attention and patch features.
% Included at the end of the appendix of main.tex via \input{appendix_visualization}.
% Figures generated by probes/visualize_cls_attention.py, probes/visualize_pca_features.py,
% and probes/visualize_attention_dynamics.py
% (SiamJEPA repo, vit-large branch).

\section{Visualization of \texttt{[CLS]} attention and patch features}
\label{sec:visualization}

To complement the quantitative analyses, we visualize what the encoders learn under different settings and how this develops during training. Table~\ref{tab:vis-models} lists the encoders; all are ViT-B/16, and for every SiamJEPA variant we use the EMA/teacher encoder. As a reference we include the official DINO ViT-B/16 checkpoint~\citep{caron2021emerging}.

\paragraph{Setup.}
Images are ImageNet-1K validation images, one drawn at random (fixed seed) from each of eight classes that contain a salient object (golden retriever, tabby cat, goldfinch, African elephant, goldfish, tree frog, airliner, mountain bike); they are resized and center-cropped to $448\times448$, and the positional embeddings are interpolated from the $14\times14$ to the $28\times28$ patch grid. \emph{\texttt{[CLS]} attention}: attention weights of the \texttt{[CLS]} query over the patch tokens in the last Transformer block, averaged over the 12 heads and min--max normalized per map. \emph{Patch-feature PCA}: a per-image 3-component PCA of the final-layer (post-LayerNorm) patch tokens mapped to RGB (per-component min--max scaling after clipping the 1st/99th percentiles); the components of each encoder are matched to those of the first column by the Hungarian algorithm on their absolute correlation (with sign flips). \emph{Training dynamics}: the \texttt{[CLS]} attention of the checkpoints saved during pretraining, shown next to the training curve of $\mathcal{L}_{\mathrm{sim1}}=\max(\mathrm{KL},\,0.1)$, i.e., the KL term after the free-bit threshold.

\begin{table}[h]
\centering
\small
\caption{Encoders analyzed in this section (ViT-B/16, final checkpoints, EMA/teacher branch; all SiamJEPA variants use the free-bit threshold 0.1). IN-1k: EMA/teacher linear-probe top-1 accuracy at the final checkpoint (cf.\ Figure~\ref{fig:kl-wd} and Table~\ref{tab:position-decodability}); the DINO number is the one reported by \citet{caron2021emerging}.}
\label{tab:vis-models}
\begin{tabular}{lcccc c}
\toprule
Encoder & $\lambda_{\mathrm{KL}}$ & wd & mask ratio & RST & IN-1k top-1 (\%) \\
\midrule
JEPA-like & $10^{-5}$ & 0.05 & 0.75 & -- & 61.55 \\
JEPA-like & $10^{-5}$ & 0.1 & 0.75 & -- & 68.30 \\
SiamJEPA & 0.01 & 0.05 & 0.75 & -- & 64.26 \\
SiamJEPA & 0.01 & 0.1 & 0.75 & -- & 65.18 \\
SiamJEPA & 0.03 & 0.1 & 0.75 & -- & 63.56 \\
SiamJEPA, mask 0.85 & 0.01 & 0.1 & 0.85 & -- & 69.04 \\
RST (400 epochs) & 0.01 & 0.1 & 0.75 & \checkmark & 71.13 \\
RST$\rightarrow$SiamJEPA (200+250 epochs) & 0.01 & 0.1 & 0.75 & first 200 epochs & 73.25 \\
DINO (official) & -- & -- & -- & -- & 78.2 \\
\bottomrule
\end{tabular}
\end{table}

\begin{figure}[h]
\centering
\includegraphics[width=0.72\linewidth]{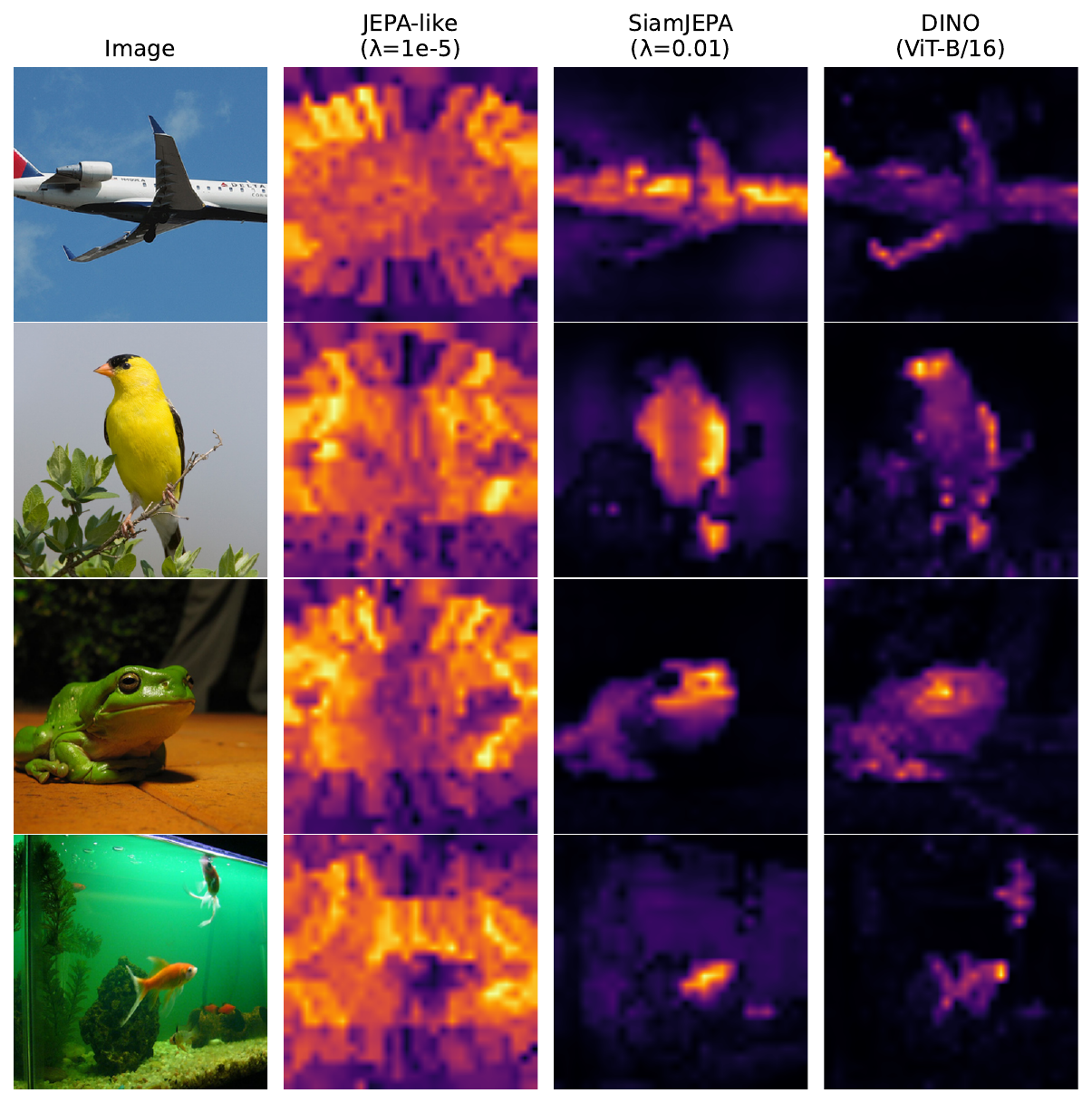}
\caption{\texttt{[CLS]}$\to$patch attention of the last block (head-averaged). JEPA-like and SiamJEPA are trained with identical settings (400 epochs, mask ratio 0.75, weight decay 0.1) except for the Siamese KL weight $\lambda_{\mathrm{KL}}$. With the Siamese term, the attention localizes the foreground object, similarly to DINO. Four representative images are shown.}
\label{fig:vis-cls-compact}
\end{figure}

\paragraph{Siamese consistency yields DINO-like \texttt{[CLS]} attention.}
For the JEPA-like encoders, the \texttt{[CLS]} attention is tied to position rather than content: it forms a blob around the image center or covers fixed image regions irrespective of the object (Figure~\ref{fig:vis-cls-compact}). With $\lambda_{\mathrm{KL}}=0.01$ and weight decay $0.1$, the attention instead concentrates on the foreground object, the airliner, the goldfinch, the frog, and even the single small goldfish in the aquarium, and different heads attend to different object parts (not shown), closely resembling DINO. Since this encoder and the JEPA-like encoder with weight decay $0.1$ differ only in $\lambda_{\mathrm{KL}}$, we attribute the difference to the Siamese term, which aligns the global representations of two views with disjoint visible patches. Object-centric \texttt{[CLS]} attention has so far been characteristic of methods that match \texttt{[CLS]} representations across views, such as DINO, rather than of reconstruction- or prediction-based methods; our observation suggests that Siamese student encoders bring this inductive bias into the JEPA framework. The object-centric encoder nevertheless retains positional information ($93.5\%$ position decodability for the EMA/teacher, Table~\ref{tab:position-decodability}).

\begin{figure}[h]
\centering
\includegraphics[width=\linewidth]{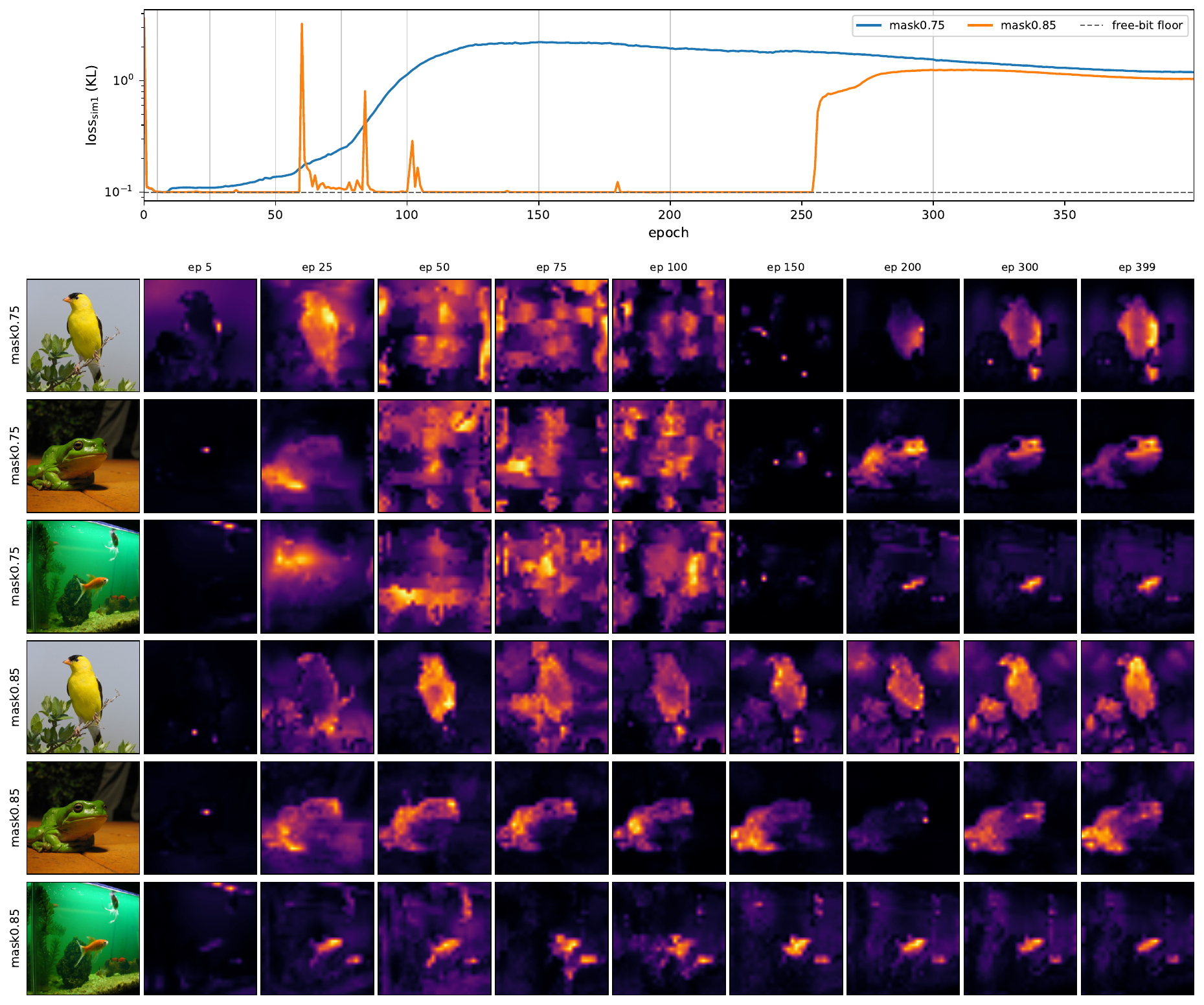}
\caption{Training dynamics for mask ratio 0.75 vs.\ 0.85 ($\lambda_{\mathrm{KL}}=0.01$, weight decay 0.1, no RST). Top: $\mathcal{L}_{\mathrm{sim1}}$ (KL after the free-bit threshold; dashed line) with the visualized epochs marked. Bottom: \texttt{[CLS]} attention of the EMA/teacher encoder at those epochs. With mask ratio 0.85, the attention is object-centric from about epoch 25 onward, whereas with 0.75 it becomes object-centric only around epoch 200.}
\label{fig:vis-dyn-mask}
\end{figure}

\paragraph{A higher mask ratio makes object-centric attention emerge early.}
Figure~\ref{fig:vis-dyn-mask} follows the \texttt{[CLS]} attention during pretraining. With mask ratio 0.75, the attention is disorganized during the first 150 epochs and becomes object-centric only around epoch 200, after which it sharpens until the end of training. With mask ratio 0.85, the attention already localizes the object at about epoch 25 and stays object-centric for the rest of training. This encoder also reaches the highest linear-probe accuracy among the models trained without RST ($69.04\%$), so object-centric attention and good linear separability co-occur here. It may also explain the robustness of the mask-0.85 encoder to occlusion, which, as discussed in Appendix~\ref{sec:per-token-semantics}, cannot be attributed to more class information per token.

\begin{figure}[h]
\centering
\includegraphics[width=\linewidth]{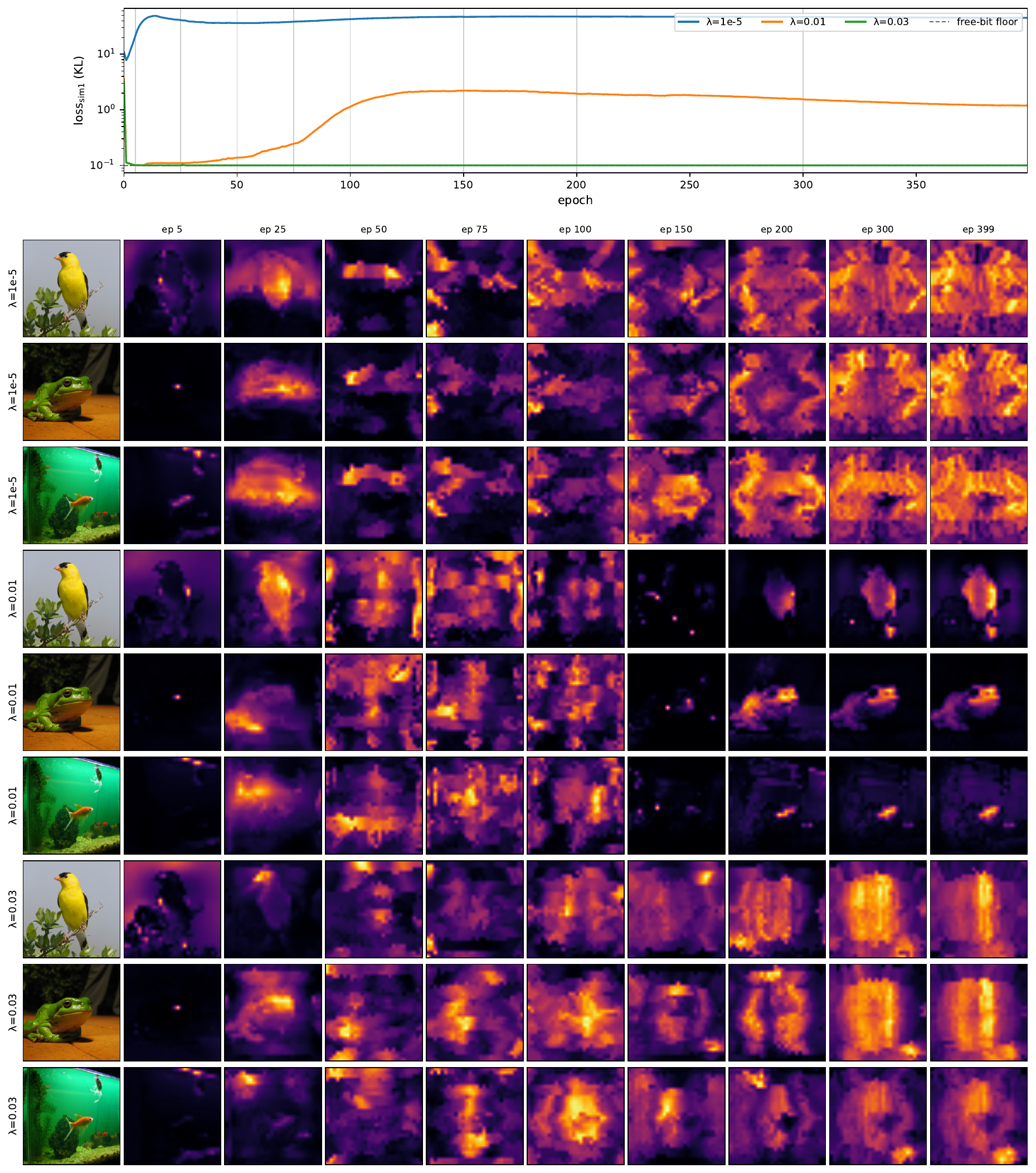}
\caption{Training dynamics for $\lambda_{\mathrm{KL}}\in\{10^{-5},0.01,0.03\}$ (mask ratio 0.75, weight decay 0.1, no RST); same layout as Figure~\ref{fig:vis-dyn-mask}.}
\label{fig:vis-dyn-lambda}
\end{figure}

\paragraph{Training dynamics across KL weights.}
Figure~\ref{fig:vis-dyn-lambda} compares three KL weights. For $\lambda_{\mathrm{KL}}=10^{-5}$, the KL term stays around 40--50 throughout training and the attention settles into fixed positional patterns. For $\lambda_{\mathrm{KL}}=0.03$, the KL term is pushed below the free-bit threshold early and stays there for all 400 epochs; since $\max(\mathrm{KL},0.1)$ has zero gradient in this regime, the Siamese term is effectively inactive, and the attention never becomes object-centric. For $\lambda_{\mathrm{KL}}=0.01$, the KL term stays close to the threshold for the first 70 epochs, rises steeply between epochs 75 and 120 to a peak of about 2.2 around epoch 150, and then slowly decreases to about 1.2; the attention is disorganized during this rise and becomes object-centric only afterwards, around epoch 200. Activity of the KL term alone, however, does not determine the outcome: with mask ratio 0.85 the KL term stays at the threshold until about epoch 250, apart from brief spikes around epochs 60, 85, and 100, yet the attention is object-centric from early on (Figure~\ref{fig:vis-dyn-mask}). Disentangling the roles of the mask ratio and of the Siamese term would require, e.g., a JEPA-like encoder with mask ratio 0.85, which we leave for future work. In all runs, the attention maps at epoch 25 show a transient blob around the object before the runs diverge.

\begin{figure}[p]
\centering
\includegraphics[width=\linewidth]{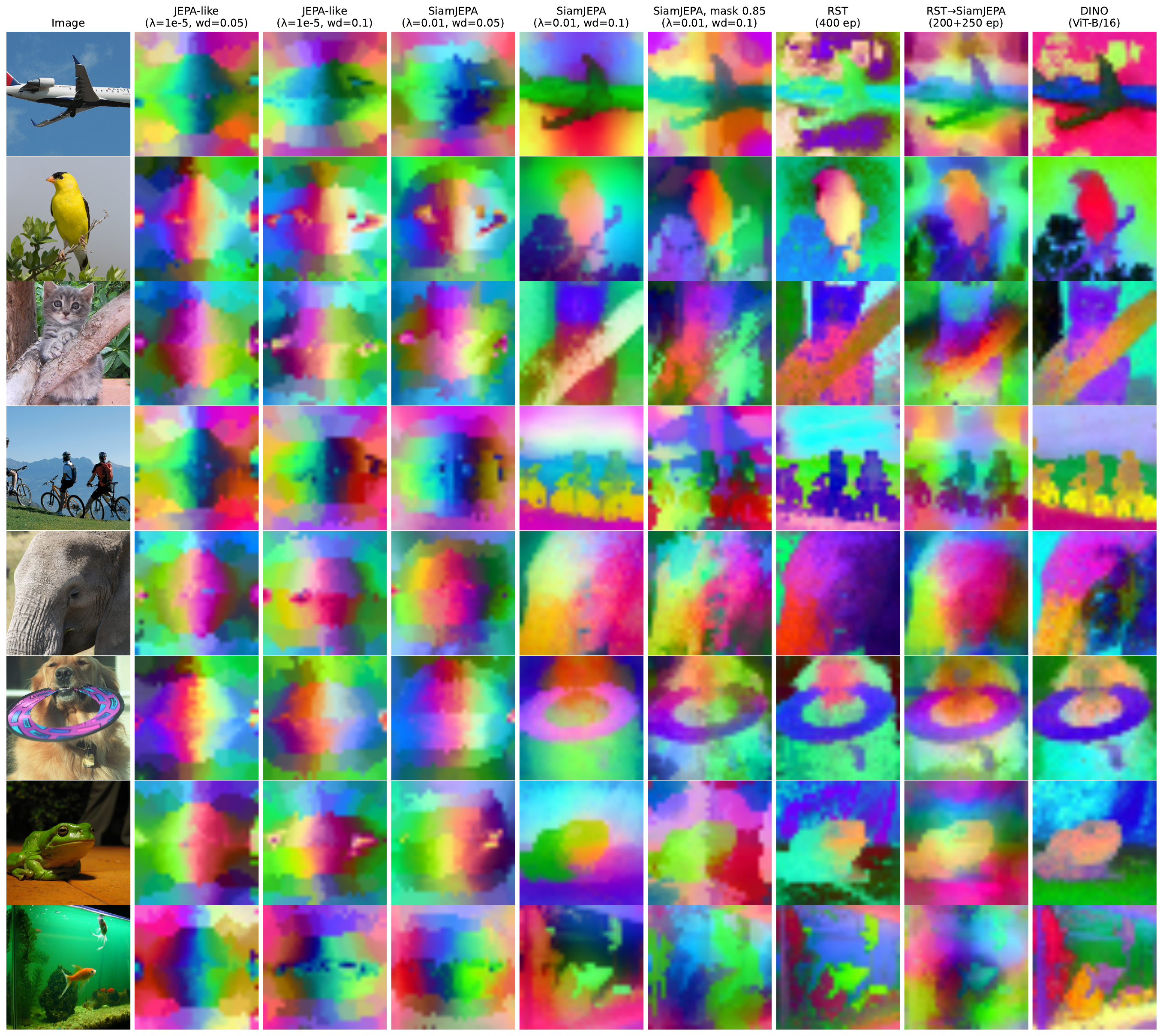}
\caption{Per-image PCA of final-layer patch features mapped to RGB (components matched across columns). Weak Siamese regularization ($\lambda_{\mathrm{KL}}=10^{-5}$) or the default weight decay (0.05) yields position-dominated components, whereas $\lambda_{\mathrm{KL}}=0.01$ with weight decay 0.1 (both mask ratios), RST, and the curriculum yield object-centric features comparable to DINO.}
\label{fig:vis-pca}
\end{figure}

\paragraph{Weak Siamese regularization: position-dominated patch features.}
The patch-feature PCA shows the same picture as the attention (Figure~\ref{fig:vis-pca}). For the JEPA-like encoders and for SiamJEPA with the default weight decay of 0.05, the leading principal components vary with the spatial position rather than with the image content: the same radial and left--right pattern appears in every image. This is consistent with the per-token analysis in Appendix~\ref{sec:per-token-semantics}, where a single patch token of the $\lambda_{\mathrm{KL}}=10^{-5}$ encoder is almost uninformative for the linear probe ($1.8\%$) despite $68.2\%$ for the 196-token average, i.e., the tokens carry large position-specific components that cancel only in the average. With $\lambda_{\mathrm{KL}}=0.01$ and weight decay 0.1 (mask ratio 0.75 or 0.85), the PCA instead separates objects and their parts from the background (e.g., the goldfinch from the branch and the sky, the cat from the tree branch, the riders from the sky and the grass), comparably to DINO.

\begin{figure}[h]
\centering
\includegraphics[width=\linewidth]{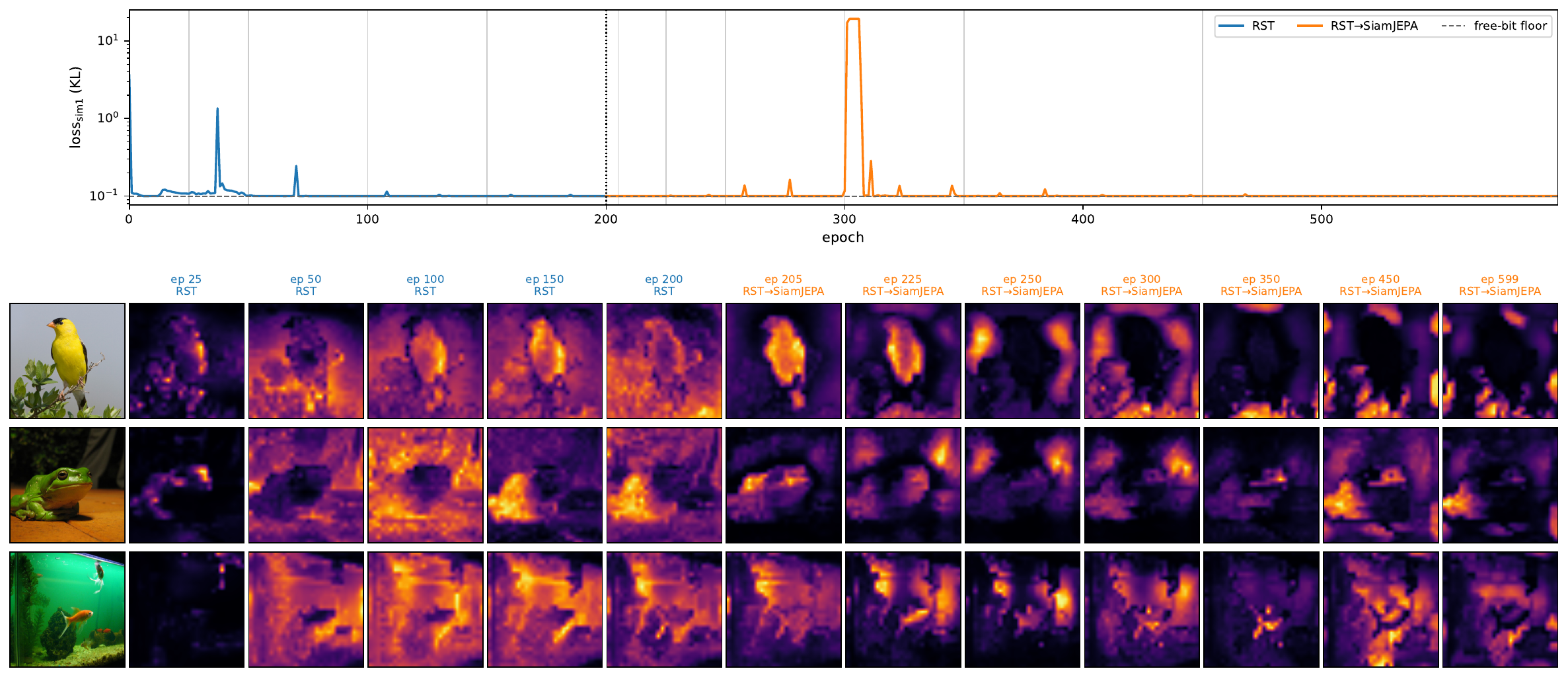}
\caption{Training dynamics along the RST-based curriculum (RST for 200 epochs, then standard SiamJEPA training initialized from the epoch-200 RST checkpoint), plotted over the total number of epochs; the dotted vertical line marks the switch. Epoch 450 is the checkpoint with $73.25\%$ linear-probe accuracy.}
\label{fig:vis-dyn-curriculum}
\end{figure}

\paragraph{RST and the curriculum.}
RST produces object-centric patch features (Figure~\ref{fig:vis-pca}), in line with the loss of positional information in Table~\ref{tab:position-decodability}, but its \texttt{[CLS]} attention is broad and extends over the background. Along the curriculum (Figure~\ref{fig:vis-dyn-curriculum}), the KL term stays at the free-bit threshold in both stages apart from brief spikes. Right after the switch (total epochs 205--225), the attention becomes sharply object-centric; from about epoch 250 onward, it shifts towards the surroundings of the object and the background. At epoch 450, the most accurate checkpoint, the \texttt{[CLS]} attention is thus not object-centric, even though the patch features still group the objects (Figure~\ref{fig:vis-pca}). Object-centric \texttt{[CLS]} attention is therefore neither necessary nor sufficient for high linear-probe accuracy in our setting.

\paragraph{Limitations.}
These analyses are qualitative and based on a single training run per setting and a small number of images. The relation between object-centric \texttt{[CLS]} attention and linear-probe accuracy is not monotonic: the mask-0.85 encoder combines both, SiamJEPA with mask ratio 0.75 localizes objects but is less accurate than the JEPA-like encoder with the same weight decay ($65.18\%$ vs.\ $68.30\%$), and the most accurate curriculum checkpoint does not show object-centric \texttt{[CLS]} attention. Quantifying the localization (e.g., by attention-based object segmentation), evaluating it on dense prediction tasks, and the missing JEPA-like control with mask ratio 0.85 are left for future work.

\end{document}